\pdfoutput=1

\documentclass[11pt]{article}

\usepackage[preprint]{acl}

\usepackage{times}
\usepackage{latexsym}

\usepackage[T1]{fontenc}

\usepackage[utf8]{inputenc}

\usepackage{microtype}

\usepackage{inconsolata}

\usepackage{graphicx}

\usepackage{booktabs}
\usepackage{multirow}
\usepackage{tipa}
\usepackage{subcaption}

\title{Fast, Not Fancy: Rethinking G2P with Rich Data and Rule-Based Models}

\author{
  Mahta Fetrat Qharabagh, Zahra Dehghanian, Hamid R. Rabiee \\
  Dep. of Computer Engineering, Sharif University of Technology \\
  \texttt{m.fetrat@sharif.edu}, 
  \texttt{zahra.dehghanian97@sharif.edu}, 
  \texttt{rabiee@sharif.edu}
}

\begin{document}
\maketitle
\begin{abstract}
Homograph disambiguation remains a significant challenge in grapheme-to-phoneme (G2P) conversion, especially for low-resource languages. This challenge is twofold: (1) creating balanced and comprehensive homograph datasets is labor-intensive and costly, and (2) specific disambiguation strategies introduce additional latency, making them unsuitable for real-time applications such as screen readers and other accessibility tools. In this paper, we address both issues. First, we propose a semi-automated pipeline for constructing homograph-focused datasets, introduce the HomoRich dataset generated through this pipeline, and demonstrate its effectiveness by applying it to enhance a state-of-the-art deep learning-based G2P system for Persian. Second, we advocate for a paradigm shift—utilizing rich offline datasets to inform the development of fast, rule-based methods suitable for latency-sensitive accessibility applications like screen readers. To this end, we improve one of the most well-known rule-based G2P systems, eSpeak, into a fast homograph-aware version, HomoFast eSpeak. Our results show an approximate 30\% improvement in homograph disambiguation accuracy for the deep learning-based and eSpeak systems.
\end{abstract}

\section{Introduction}

Grapheme-to-phoneme (G2P) conversion is a crucial step in many fast text-to-speech (TTS) models \citeyearpar{ren2020fastspeech}. It refers to the task of converting a given written text into its corresponding sequence of phonemes—how it is pronounced. There are several formats for representing phoneme sequences, one of the most widely used being the International Phonetic Alphabet (IPA) \citeyearpar{international1999handbook}. As an example, the phoneme sequence for the sentence \textit{"I will read it"} is \textipa{/aI wIl ri:d It/} in IPA format.

The complexity of G2P conversion varies by language. Some languages like Turkish and Spanish are highly phonetic, meaning a near one-to-one correspondence between spelling and pronunciation \citeyearpar{kocsaner2013improving, delattre1945spanish}. In contrast, in many other languages, such as Persian, G2P is more complex due to exceptions and rules that depend on context \citeyearpar{qharabagh2025llm}.

One such challenge is handling homographs—words spelled the same but pronounced differently depending on context. For example, the word \textit{"read"} is pronounced \textipa{/rEd/} in the past tense (\textit{"I read this book yesterday"}) and \textipa{/ri:d/} in the present tense (\textit{"I read the book every night"}).

Unfortunately, sentence-level G2P datasets are extremely scarce in low-resource languages. This scarcity stems from the fact that phonemization is a time-consuming and costly process that requires expert annotators. Homograph-specific datasets are even rarer, as they depend on source corpora that must meet strict conditions: they should contain a wide range of homographs and provide a balanced number of examples for each pronunciation. Without this balance, the resulting models will fail to learn some homographs and tend to default to the more frequent pronunciation in ambiguous cases.

Beyond data scarcity, there is also a methodological challenge in G2P conversion. The two primary approaches are rule-based methods \citeyearpar{silva2012rule,alayiaboozar2019word,riahi2012semi} and neural models \citeyearpar{ploujnikov2024towards,vrezavckova2024t5g2p,gao2024unsupervised}. While neural methods have gained popularity due to their flexibility and learning capacity, they often suffer from high inference latency, making them unsuitable for real-time applications such as screen readers that serve accessibility needs. This motivates a renewed focus on rule-based approaches, aiming to improve their accuracy while preserving their inherent speed.

This work proposes a practical approach for generating a rich and balanced homograph dataset. We demonstrate that such a dataset not only boosts the homograph disambiguation accuracy of neural G2P models but also significantly enhances the performance of rule-based systems. Specifically, we show that by incorporating a simple, fast statistical method that leverages the proposed dataset, rule-based models can be equipped with context understanding, leading to improved handling of homographs without sacrificing speed.

Our key contributions are as follows:
\begin{itemize}
    \item We propose a practical and cost-efficient recipe for constructing rich and balanced homograph datasets in low-resource languages by leveraging LLMs for G2P annotation and homograph sample generation.

    \item We release HomoRich, the first and largest Persian homograph dataset, and demonstrate its effectiveness by improving the homograph disambiguation accuracy of a state-of-the-art neural G2P model by 29.72\%.

    \item We introduce a lightweight statistical method that enhances G2P systems for homograph disambiguation, using datasets generated by our proposed approach.

    \item We integrate this method into the open-source eSpeak engine, resulting in HomoFast eSpeak, a variant that achieves a 30.66\% improvement in homograph disambiguation without compromising real-time performance.
\end{itemize}

\section{Related Works}
\label{sec:related-works}

In this section, we review homograph disambiguation from two perspectives: the common methods and the datasets developed to address challenges in low-resource settings like Persian.

\begin{table*}
\centering
\begin{tabular}{lcccccc}
\toprule
\multirow{2}{*}{\textbf{Title}} & \textbf{Sample} & \textbf{Sample} & \textbf{Hom.} & \textbf{Hom.} & \multirow{2}{*}{\textbf{Availablity}} & \multirow{2}{*}{\textbf{License}} \\
 & \textbf{Type} & \textbf{Count} & \textbf{Curated} & \textbf{Count} &  &  \\ \midrule
\citeyearpar{riahi2012semi} Semi-sup H.D. & Sent. & -- & Yes & {\color{red!60!black}2} & {\color{red!60!black}Not avail.} & {\color{red!60!black}N.A.} \\
\multirow{2}{*}{\citeyearpar{moghadaszadeh2024avashog2p} AvashoG2P } & Sent. & -- & Yes & {\color{red!60!black}54} & {\color{red!60!black}Not avail.} & {\color{red!60!black}N.A.} \\
 & {\color{red!60!black}Word} & 12,000 & {\color{red!60!black}No}
 & {\color{red!60!black}--} & Available & {\color{red!60!black}N.A.} \\
 \citeyearpar{rezaei2022multi} Multi-Module G2P  & Sent. & 42,540 & {\color{red!60!black}No} & {\color{red!60!black}--} & {\color{red!60!black}Not avail.} & {\color{red!60!black}N.A.} \\
\citeyearpar{rahmati2024ge2pe} GE2PE  & Sent. & 5,376,670 & {\color{red!60!black}No} & {\color{red!60!black}--} & Available & MIT \\
\textbf{HomoRich (Ours)} & \textbf{Sent.} & \textbf{528,891} & \textbf{Yes} & \textbf{285} & \textbf{Available} & \textbf{CC0-1.0} \\
\bottomrule
\end{tabular}
\caption{Persian Homograph Datasets. \textit{Hom. Count} shows the number of homographs covered in the dataset and \textit{Hom. Curated} indicates if homograph samples were deliberately inserted or naturally occurring in a regular corpus.}
\label{tab:comparison}
\end{table*}

\subsection{Approaches}

There are multiple approaches to addressing the homograph challenge, including neural and rule-based methods, various machine learning algorithms, hybrid techniques, and the use of large language models (LLMs). We briefly highlight only the works most relevant to our approach. A more comprehensive review is provided in Appendix~\ref{appendix:extended-related-work}.


\paragraph{Rule-based approaches} have been widely explored for homograph disambiguation across various languages. These methods often rely on morphosyntactic patterns, lexical cues, and contextual heuristics rather than deep semantic inference. For instance, \citet{silva2012rule} and \citet{alayiaboozar2019word} utilized hand-crafted linguistic rules derived from syntactic and morphological features in Brazilian Portuguese and Persian, respectively. \citet{hearst1991noun} introduced a system based on shallow syntactic patterns and lexical co-occurrences in local contexts, while \citet{yarowsky1997homograph} developed data-driven decision lists using log-likelihood-ranked contextual patterns. \citet{riahi2012semi} further extended these ideas by integrating rule-based decision lists into a tri-training framework.


\paragraph{Neural approaches} have been widely adopted for homograph disambiguation and G2P conversion across languages, leveraging contextual embeddings, sequence modeling, and attention mechanisms. \citet{nicolis2021homograph} and \citet{seale2021label} utilized pretrained language models like BERT, ALBERT, and XLNet to extract contextual word embeddings and fine-tune token classifiers or logistic regressors for English homographs. SoundChoice, proposed by \citet{ploujnikov2024towards}, employed a hybrid RNN-attention model with BERT embeddings and curriculum learning to predict phonemes in context. Similarly, \citet{nanni2023disambiguating} adapted SoundChoice for Italian, integrating ChatGPT-generated data. \citet{vrezavckova2024homograph,vrezavckova2024t5g2p} adopted the T5 transformer for multilingual G2P, bypassing rule-based post-processing by modeling cross-word effects. 
\citet{comini2025lightweight} combined GRUs, transformers, and knowledge distillation for efficient G2P in low-resource settings. \citet{gao2024unsupervised} enhanced multilingual phonetic recognition using self-supervised learning models (e.g., wav2vec2, HuBERT) and synthetic data.

\paragraph{LLM-based approaches} are increasingly demonstrating the potential of LLMs in G2P conversion.  \citet{suvarna2024phonologybench} were the first to benchmark models like GPT-4 and Claude-3 on phonological tasks, including G2P, and found that while promising, they still lag behind traditional models in accuracy. \citet{han2024improving} leveraged GPT-4’s in-context retrieval to map homographs to dictionary pronunciations, combining automated generation with manual refinement for accuracy. Similarly, in our previous work \cite{qharabagh2025llm}, we applied LLMs to Persian G2P conversion through advanced prompting, achieving state-of-the-art results on custom datasets without model fine-tuning.

\subsection{Datasets}

Several studies have proposed various methods to address data scarcity in G2P for low-resource languages such as Persian. A comprehensive review of these studies is provided in Appendix~\ref{sec:appendix-datasets-review}; however, here we summarize only the most relevant features of the datasets in Table~\ref{tab:comparison}. As shown, all of the referenced datasets are either not homograph-specific, not sentence-level, or not publicly available. This highlights a critical gap in homograph data for Persian—and likely for many other low-resource languages—which has resulted in the lack of G2P systems that outperform random chance in homograph disambiguation.

\section{Methodology}
Developing an effective G2P model requires both high-quality data and the tools to make use of it. This section outlines our data generation process and how we leveraged it to improve G2P models.

\subsection{Data Preparation}
The scarcity of homograph data arises from two main challenges. First, assembling a high-quality text corpus that provides broad and balanced coverage of homographs across diverse contexts is difficult. Second, phonemizing a text corpus is both time-consuming and costly, as it requires trained experts with linguistic knowledge. In this paper, we present a practical approach for collecting such data in a low-resource language like Persian and demonstrate its effectiveness in the next section.

To tackle the first challenge, we started with KaamelDict \cite{fetrat_kaameldict}, the most extensive Persian G2P dictionary introduced in \citealp{qharabagh2025llm}. We filtered for words with multiple valid pronunciations to identify potential homographs. Then, through manual review, we excluded words that either (1) had multiple commonly accepted pronunciations needing no disambiguation, or (2) included archaic, poetic, or rarely used forms. From this, we selected a list of 285 homograph words that were both comprehensive and practically relevant.

The next task was to generate a diverse and balanced set of sentences for each homograph, covering different usage contexts and ensuring equal representation of all pronunciations.
To automate this, we experimented with prompting LLMs to generate sentences for each pronunciation or meaning. However, the results were often skewed toward the dominant pronunciation, even with explicit instructions. We found that embedding the homograph in a full sentence that implied its intended meaning significantly improved accuracy.

\begin{figure*}
    \includegraphics[width=0.8\textwidth]{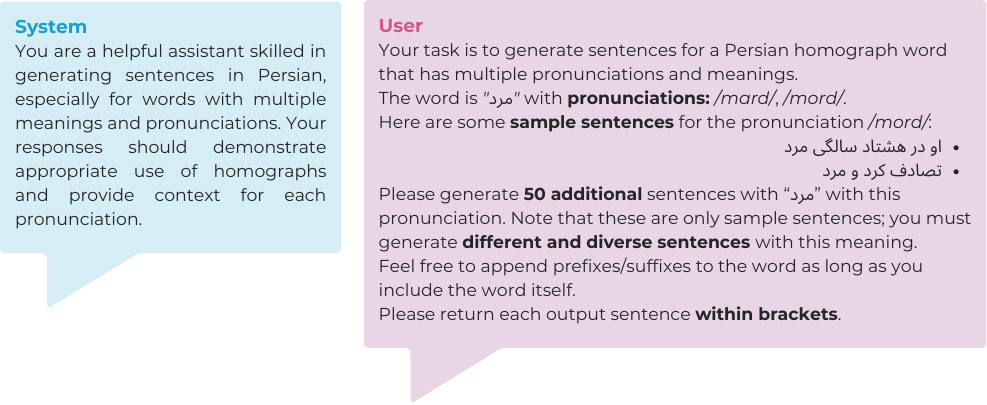}
  \centering
  \caption{Prompt for generating homograph sentences.}
  \label{fig:hom-gen-prompt}
\end{figure*}

As a result, we adopted a hybrid approach, combining manual and LLM-generated sentences. We first shared a list of selected homographs with about 200 native speakers, asking each to write five contextually varied sentences for every pronunciation. We then used some of these human-written examples as few-shot prompts to guide LLM-based sentence generation (see Figure~\ref{fig:hom-gen-prompt}).

To further enhance the dataset and support downstream TTS and G2P tasks, we integrated sentences from three widely used Persian corpora: ManaTTS \citeyearpar{qharabagh2025manatts}, GPTInformal \citeyearpar{fetrat_gptinformal_persian}, and CommonVoice \citeyearpar{ardila2019common}. These additions were meant to improve overall G2P accuracy—particularly phoneme error rate (PER)—and enrich the corpus with phoneme-annotated examples from diverse registers.

To address the second challenge—phonemization—we leveraged our prior work on LLM-powered G2P conversion \citeyearpar{qharabagh2025llm}. In that study, we demonstrated that LLMs can assist in labeling graphemes with their phonemes, thanks to their phonetic knowledge and contextual understanding, which is particularly helpful in disambiguating homographs. We introduced several techniques to enhance LLM performance in G2P tasks without requiring any training, benchmarking state-of-the-art models to guide future dataset generation.

\begin{figure*}
    \includegraphics[width=0.9\textwidth]{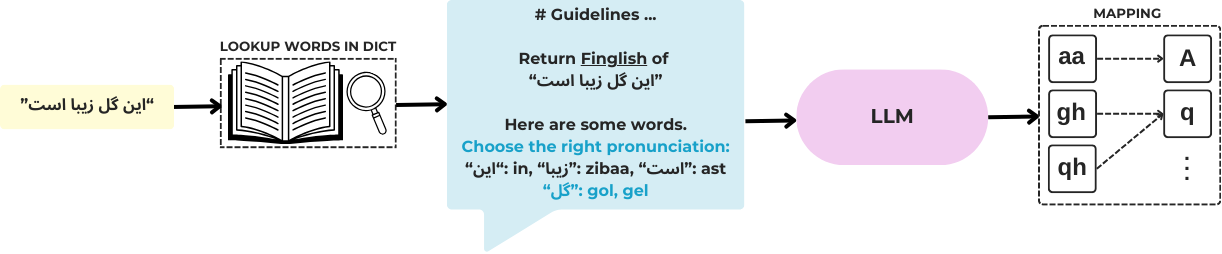}
  \centering
  \caption{LLM-powered G2P workflow \citeyearpar{qharabagh2025llm}}
  \label{fig:llm-g2p}
\end{figure*}

We use the most effective method from that study to phonemize our corpus. It prompts the model with Finglish—a more accessible but slightly ambiguous phonemic representation of Persian—instead of the less common IPA format. The method combines in-context learning, few-shot examples, hints from a G2P dictionary, and a final mapping step to produce the target phoneme format (see Figure~\ref{fig:llm-g2p}). To balance cost, availability, and quality, we use GPT-4o \citeyearpar{hurst2024gpt} as the LLM, which achieved a Phoneme Error Rate (PER) of 6.43\% and a homograph disambiguation accuracy of 64\%, outperforming many existing Persian G2P systems (see Section~\ref{results} for details).

\begin{figure}[t]
    \includegraphics[width=\columnwidth]{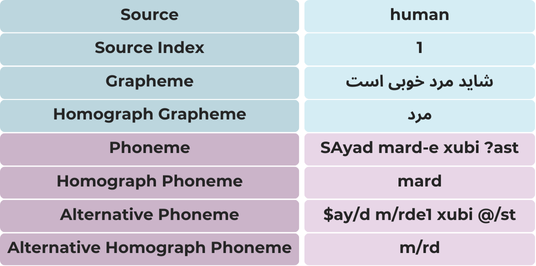}
  \centering
  \caption{Dataset structure with example entry.}
  \label{fig:dataset-sample}
\end{figure}

Figure~\ref{fig:dataset-sample}
summarizes the structure of the generated dataset.
For compatibility with previous work, we mapped the phonemes of all sentences to an alternative phoneme format (see Appendix~\ref{sec:mapping}).
We release our dataset, named HomoRich, under a permissive CC0-1 license, making it freely available for both academic and commercial use.\footnote{The HomoRich dataset is available at \url{https://huggingface.co/datasets/MahtaFetrat/HomoRich-G2P-Persian}.}

\subsubsection{Data Statistics}

\begin{figure}[t]
    \includegraphics[width=\columnwidth]{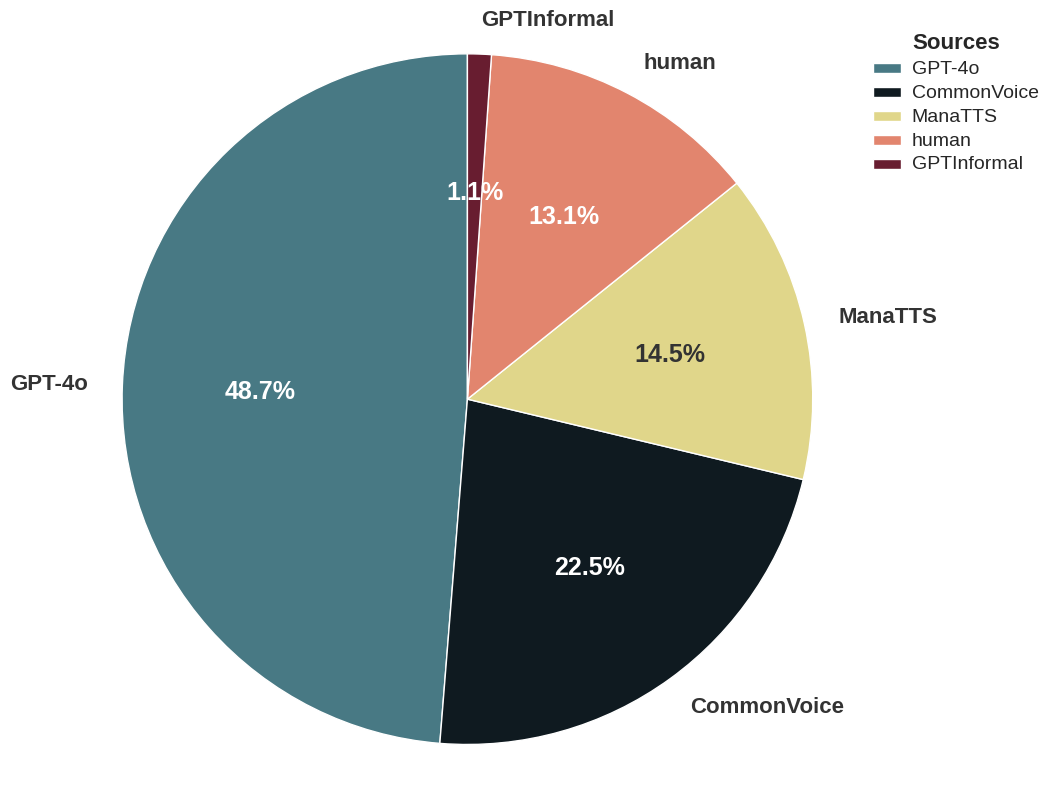}
  \centering
  \caption{Data source distribution in HomoRich dataset.}
  \label{fig:source-pie}
\end{figure}

The HomoRich dataset, generated using our proposed recipe, contains 528,891 annotated Persian sentences. As mentioned, it consists of both homograph-focused and general-purpose G2P data collected from multiple sources. Figure~\ref{fig:source-pie} and Table~\ref{tab:sources} illustrate the composition of the dataset.

\begin{table}
  \centering
  \begin{tabular}{lr}
        \toprule
        \textbf{Source} & \textbf{Count} \\
        \midrule
        GPT-4o          & 257,915 \\
        CommonVoice     & 118,983 \\
        ManaTTS         & 76,561 \\
        human           & 69,560 \\
        GPTInformal     & 5,872 \\
        \midrule
        \textbf{Homograph Samples} & \textbf{327,475} \\
        \textbf{(Human + GPT-4o)} & \\
        \midrule
        \textbf{Total} & \textbf{528,891} \\
        \bottomrule
    \end{tabular}
  \caption{The source for different parts of the HomoRich dataset.}
  \label{tab:sources}
\end{table}

To ensure diversity, both human annotators and language models were instructed to generate data across a wide range of contexts. The dataset comprises 75,715 unique words, and the distribution of sentence lengths is shown in Appendix Figure~\ref{fig:word-count-hist}.

\begin{figure}[t]
    \centering
    \includegraphics[width=\columnwidth]{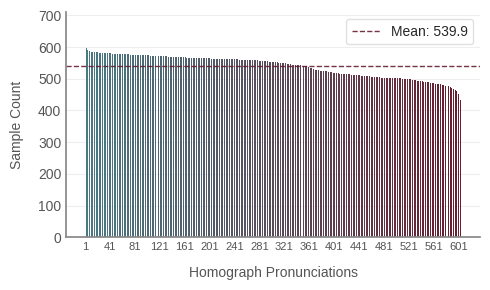}
    \caption{Sample counts per pronunciation.}
    \label{fig:phoneme-hist}
\end{figure}

The HomoRich dataset includes 285 homograph words, each associated with multiple pronunciations: 257 have two variants, 21 have three, and 7 have four.
On average, each homograph appears in over 1,000 distinct sentence contexts. To avoid bias toward more frequent pronunciations, we maintained a balanced number of samples for each variant. Figure~\ref{fig:phoneme-hist} shows the pronunciation distribution, confirming the dataset's high balance.

\subsubsection{Data Augmentation}
\begin{figure*}[t]
    \centering
    \begin{subfigure}[b]{0.3\textwidth}
        \centering
        \includegraphics[width=\textwidth]{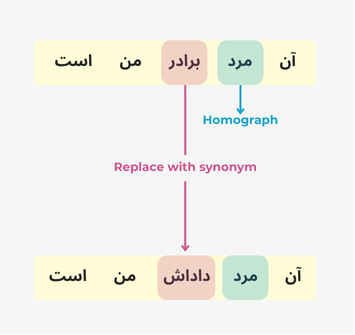}
        \caption{Synonym Replacement}
        \label{fig:aug1}
    \end{subfigure}
    \hfill
    \begin{subfigure}[b]{0.3\textwidth}
        \centering
        \includegraphics[width=\textwidth]{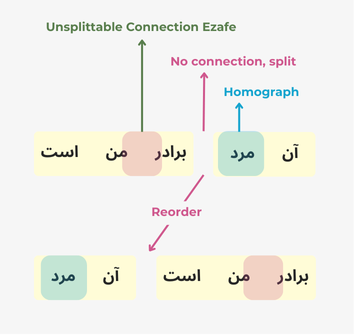}
        \caption{Reordering}
        \label{fig:aug2}
    \end{subfigure}
    \hfill
    \begin{subfigure}[b]{0.3\textwidth}
        \centering
        \includegraphics[width=\textwidth]{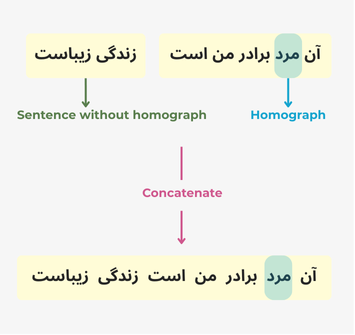}
        \caption{Concatenation}
        \label{fig:aug3}
    \end{subfigure}
    \caption{Illustration of our three data augmentation methods for homograph disambiguation.}
    \label{fig:augmentation}
\end{figure*}

To further address data scarcity—particularly in homograph disambiguation—we proposed three augmentation methods (Figure~\ref{fig:augmentation}) aimed at enhancing the model's understanding of context and increasing data diversity.

\begin{enumerate}
    \item Synonym Replacement (Figure~\ref{fig:augmentation}a): We identified the most frequently occurring words in the dataset and mapped each to a set of synonyms with equivalent meaning. For each sentence, we replaced these words with their alternatives to generate new samples.
    \item Sentence Reordering (Figure~\ref{fig:augmentation}b): In most cases, the order of context words does not affect the pronunciation of the homograph. Thus, we split sentences at random words and swapped the resulting segments, updating their corresponding phoneme sequences. However, in Persian and similar languages like Arabic, Ezafe (a phoneme that connects grammatically related words) must be preserved. We employed a POS tagger \citeyearpar{hazm} to detect Ezafe constructions and ensured no splits occurred within them.
    \item Homograph-focused Concatenation (Figure~\ref{fig:augmentation}c): we further augmented homograph samples by appending randomly selected short sentences (without homographs) to the homograph samples.
\end{enumerate}

Using combinations of these methods, we were able to scale the dataset by up to 10x, depending on the augmentation configuration.

\subsection{Proposed G2P Tools}
Having generated a large, rich, and balanced homograph dataset using the proposed method, we introduce both neural and rule-based G2P tools that build upon this data and demonstrate how this dataset can be used to enhance homograph disambiguation in each approach.

\subsubsection{Homo-GE2PE (Neural)}
\label{sec:homo-ge2pe}

As reviewed in Section~\ref{sec:related-works}, T5 has been successfully fine-tuned for G2P tasks in multiple studies \citeyearpar{vrezavckova2024t5g2p,vrezavckova2024homograph,rahmati2024ge2pe}. In a recent study \citeyearpar{rahmati2024ge2pe}, this approach resulted in GE2PE, a model achieving state-of-the-art performance in Persian G2P.
We further fine-tuned GE2PE on our dataset using a three-phase process:
\begin{enumerate}
\item Initial fine-tuning on the regular G2P subset
\item Second-phase fine-tuning on LLM-generated homograph sentences
\item Final fine-tuning on high-quality, human-authored homograph sentences
\end{enumerate}

\begin{figure*}[t]
    \centering
    \begin{subfigure}[b]{0.3\textwidth}
        \includegraphics[width=\textwidth]{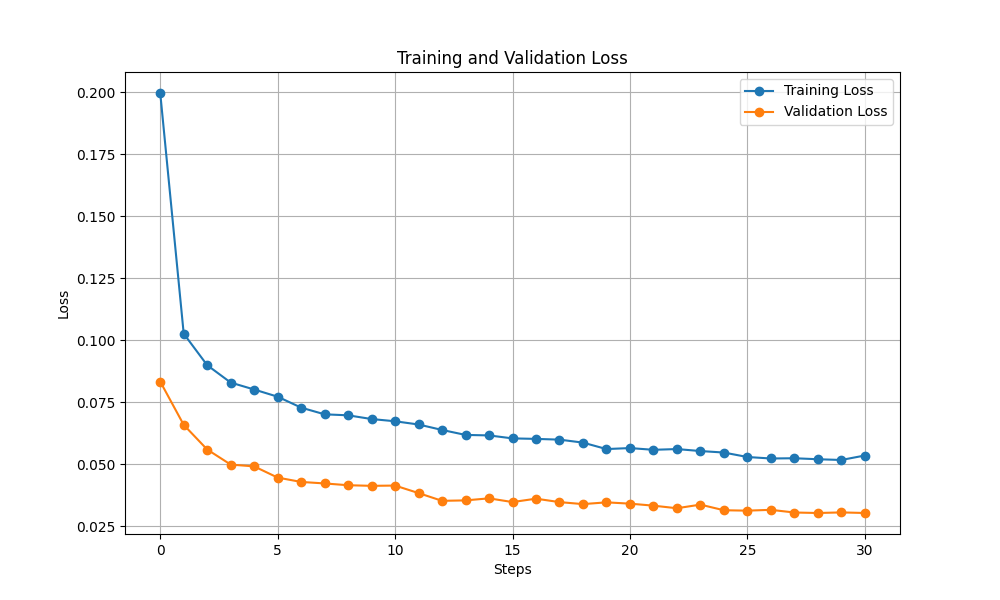}
        \caption{Phase 1 (5 epochs)}
        \label{fig:phase1}
    \end{subfigure}
    \hfill
    \begin{subfigure}[b]{0.3\textwidth}
        \includegraphics[width=\textwidth]{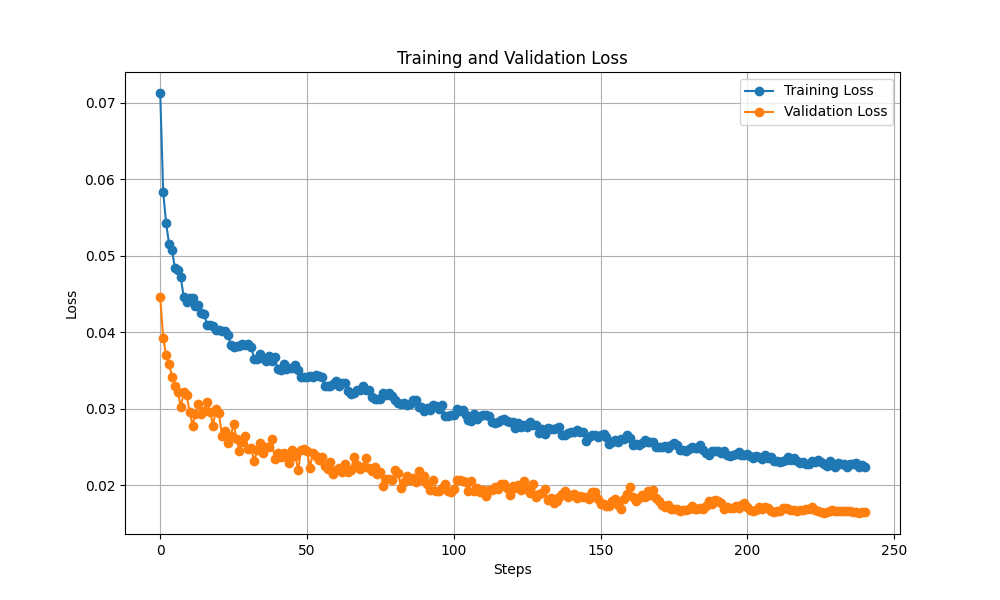}
        \caption{Phase 2 (20 epochs)}
        \label{fig:phase2}
    \end{subfigure}
    \hfill
    \begin{subfigure}[b]{0.3\textwidth}
        \includegraphics[width=\textwidth]{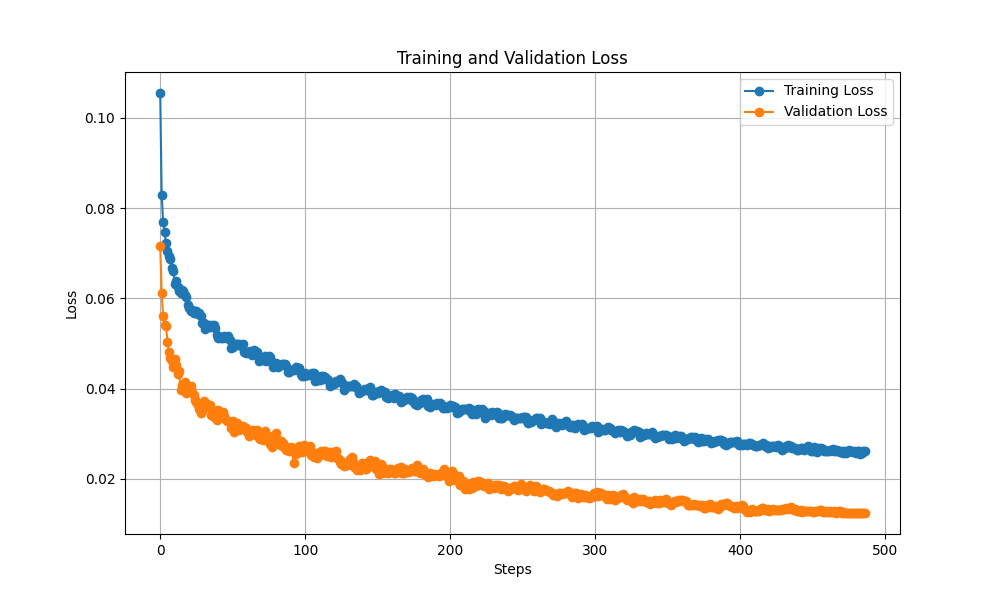}
        \caption{Phase 3 (50 epochs)}
        \label{fig:phase3}
    \end{subfigure}
    \caption{Learning curves across fine-tuning phases.}
    \label{fig:learning_curves}
\end{figure*}

We used a learning rate of 5e-4 and a batch size of 32 across all phases, with 5, 20, and 50 training epochs respectively, trained on an NVIDIA GTX TITAN X (12GB VRAM, CUDA 12.2) with Intel i7-5820K CPU. The full training process took approximately 24 hours in total. The learning curves for all phases, including training and validation metrics, are shown in Figure~\ref{fig:learning_curves}.
The resulting enhanced model, named Homo-GE2PE, is publicly available under an open license.\footnote{Complete training scripts, model files and usage instructions are available at \url{https://github.com/MahtaFetrat/Homo-GE2PE-Persian}.}

\subsubsection{HomoFast eSpeak (Rule-based)}
As discussed earlier, one of the main motivations for favoring rule-based methods in certain applications is their low latency. Neural models, while powerful, often incur high inference times, making them less suitable for real-time systems such as screen readers. In contrast, rule-based systems are extremely fast and lightweight, enabling them to operate effectively in low-latency environments. Therefore, despite the advances in neural G2P systems, it remains important to continue exploring and enhancing rule-based approaches, particularly when speed and responsiveness are critical.

However, a key limitation of rule-based systems is their difficulty in disambiguating homographs, due to their limited or nonexistent semantic or contextual understanding.
In this work, we introduce a strategy to enhance the homograph disambiguation ability of G2P systems using datasets generated by our proposed approach.
This strategy is purely statistical and does not rely on neural models or even embeddings, making it a perfect solution for improving the homograph accuracy of rule-based methods without compromising their key advantage—speed and low latency. While straightforward in design, this approach has not been explored in prior homograph disambiguation research.

The approach begins by tokenizing the sentences in our dataset, removing stopwords, and constructing a database that maps different pronunciations of homographs to lists of context words that frequently co-occur with each pronunciation.

\begin{figure*}[t]
    \centering
    \includegraphics[width=0.9\textwidth]{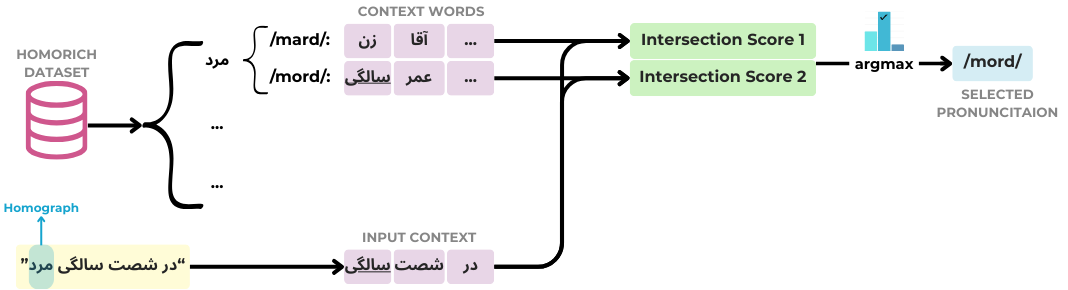}
    \caption{Overview of the proposed statistical homograph disambiguation approach.}
    \label{fig:method}
\end{figure*}

For a new sentence, we compute a weighted overlap between its context words and each pronunciation’s context list to derive a similarity score. To mitigate bias toward longer lists, we normalize each score by the length of the corresponding context list. The pronunciation with the highest normalized score is then selected as the most contextually appropriate. For a schematic overview of this method, see Figure~\ref{fig:method}.

We applied this approach to the widely used eSpeak NG project \citeyearpar{espeak}, selected for its relevance to real-world applications. eSpeak NG is a compact, open-source text-to-speech synthesizer available on Linux, Windows, Android, and other platforms. It supports over 100 languages and accents, benefiting from contributions by various linguistic communities. Notably, it has an add-on in the open-source NVDA screen reader \citeyearpar{nvaccess_website}, and its Persian G2P module is extensively used in screen readers by a large portion of the blind community in Iran \citeyearpar{nvda_speechplayer_espeak, gooshkon_espeak}.
We name the enhanced version HomoFast eSpeak, which, as shown in the following sections, demonstrated outstanding results, indicating a viable path for enhancing rule-based TTS systems in Persian.\footnote{The HomoFast eSpeak is available at \url{https://github.com/MahtaFetrat/HomoFast-eSpeak-Persian}.}


\section{Results}
\label{results}

Although several word-to-phoneme datasets exist for Persian \citeyearpar{ipa-dict,tihu-dict,IPA-Translator,wiktionary,zaya,jame,mohammadhasan,PasaOpasen,azamrabiee,zhu2022charsiu-g2p}, there was no public sentence-level dataset suitable for benchmarking homograph accuracy of G2P systems prior to our LLM-Powered G2P work \citeyearpar{qharabagh2025llm}, which introduced SentenceBench \cite{fetrat2024sentencebench}. We also adopted this dataset as the primary benchmark in our experiments.


\paragraph{Evaluation of Baseline G2P Tools:}
Table ~\ref{tab:experiment} presents the performance of previously available G2P tools on the SentenceBench benchmark.
As shown, the only two models that perform well in terms of PER are the neural GE2PE model \citeyearpar{rahmati2024ge2pe} and the rule-based eSpeak tool \citeyearpar{espeak}.
However, even these models perform worse than random when it comes to homograph disambiguation.

\begin{table*}[t]
\centering
\begin{tabular}{@{}lccc@{}}
\toprule
\textbf{Model} & \textbf{PER (\%) ↓} & \textbf{Homograph Acc. (\%) ↑} & \textbf{Avg. Inf. Time (s) ↓} \\
\midrule
\textbf{persian-phonemizer} \citeyearpar{persian-phonemizer} & 25.27 $\pm$ 0.09 & 29.25 $\pm$ 0.47 & 0.1803 $\pm$ 0.04 \\
\textbf{PersianG2P} \citeyearpar{PasaOpasen} & 15.04 $\pm$ 0.00 & 37.74 $\pm$ 0.00 & 2.1686 $\pm$ 0.10 \\
\textbf{Persian\_G2P} \citeyearpar{azamrabiee} & 35.23 $\pm$ 0.00 & 21.23 $\pm$ 0.00 & 11.1374 $\pm$ 0.56 \\
\textbf{G2P} \citeyearpar{mohammadhasan} & 19.63 $\pm$ 1.83 & 29.91 $\pm$ 0.72 & 28.0039 $\pm$ 0.42 \\
\textbf{G2P with Transformer} \citeyearpar{sajadalipour7} & 12.85 $\pm$ 0.09 & 40.00 $\pm$ 0.21 & 0.9685 $\pm$ 0.03 \\
\textbf{Epitran} \citeyearpar{Mortensen-et-al:2018} & 45.12 $\pm$ 0.00 & 0.00 $\pm$ 0.00 & $\mathbf{0.0003 \pm 0.00}$ \\
\textbf{eSpeak} \citeyearpar{espeak} & 6.92 $\pm$ 0.00 & 43.87 $\pm$ 0.00 & 0.0169 $\pm$ 0.00 \\
\textbf{GE2PE} \citeyearpar{rahmati2024ge2pe} & 4.81 $\pm$ 0.00 & 47.17 $\pm$ 0.00 & 0.4464 $\pm$ 0.03 \\
\midrule
\textbf{Homo-T5} & $\underline{4.12 \pm 0.13}$ & $\underline{76.32 \pm 0.52}$ & 0.4141 $\pm$ 0.09 \\
\textbf{HomoFast eSpeak} & 6.33 $\pm$ 0.00 & 74.53 $\pm$ 0.00 & \underline{0.0084 $\pm$ 0.00} \\
\textbf{Homo-GE2PE} & $\mathbf{3.98 \pm 0.00}$ & $\mathbf{76.89 \pm 0.00}$ & 0.4473 $\pm$ 0.02 \\
\bottomrule
\end{tabular}
\caption{Comparison of Persian G2P tools in terms of Phoneme Error Rate (PER), Homograph Accuracy, and Average Inference Time. Results are reported as mean $\pm$ standard deviation across 5 independent runs. Best results are in \textbf{bold}, and second-best are \underline{underlined}.}

\label{tab:experiment}
\end{table*}

\begin{figure*}[t]
\centering
\includegraphics[width=\textwidth]{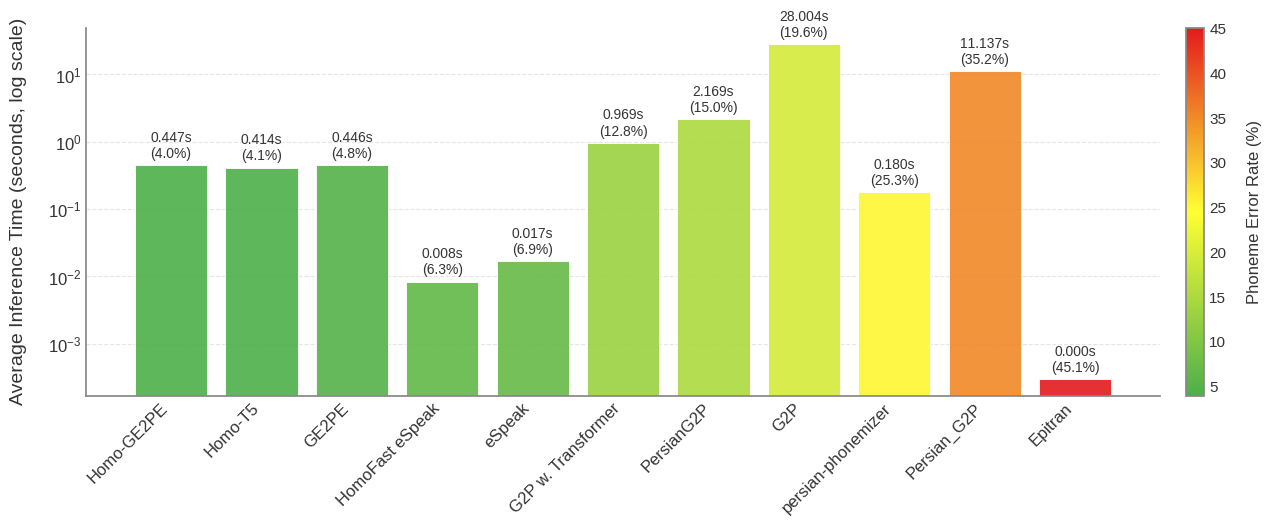} 
\caption{Inference speed and phoneme error rate (PER) of available and proposed G2P tools.}
\label{fig:speed-visualization}
\end{figure*}


\paragraph{Evaluation of the Proposed Improved G2P Tools:}
To address the challenge of homograph disambiguation in Persian G2P systems, we utilized a curated homograph dataset to enhance both neural and rule-based models. Specifically, we fine-tuned the GE2PE \citeyearpar{rahmati2024ge2pe} model and proposed a statistical disambiguation module integrated into eSpeak \citeyearpar{espeak}, resulting in two improved variants: Homo-GE2PE and HomoFast eSpeak.

As depicted in Table~\ref{tab:experiment}, our improved GE2PE model achieves a 29.72\% increase in homograph accuracy with a concurrent reduction in PER. Notably, our statistical disambiguation module—devoid of any neural components or learned embeddings—delivers the same level of homograph accuracy improvement when integrated into rule-based models, all while maintaining their inference speed. This underscores the value of high-quality data and shows that even simple statistical techniques can be highly effective when supported by strong datasets.


\paragraph{Fine-tuning T5 on Our Dataset:}
To evaluate the effectiveness of our dataset in improving both the general phoneme error rate (PER) and homograph disambiguation, we fine-tuned the base GE2PE model (T5) using only our data with the same hardware setup and training configuration as for Homo-GE2PE, referring to this variant as Homo-T5. The learning curves can be seen in Figure~\ref{fig:t5-learning-curves}. Despite our dataset being an order of magnitude smaller than the 5-million-sample synthetic dataset used in the original GE2PE study \citeyearpar{rahmati2024ge2pe}, Homo-T5 achieves competitive PER and high homograph accuracy (Table~\ref{tab:experiment}), demonstrating the quality and utility of our approach.

\paragraph{Evaluation of Inference Speed:}
Another critical factor is inference speed. While the Homo-GE2PE model outperforms HomoFast eSpeak in accuracy, it is orders of magnitude slower, making it impractical for real-time applications such as screen readers.
Figure~\ref{fig:speed-visualization} presents the speed and accuracy of all available and proposed G2P tools. All inference tests were conducted on Google Colab (CPU runtime).\footnote{Inference scripts and benchmarking code are available at \url{https://github.com/MahtaFetrat/Persian-G2P-Tools-Benchmark}.} The color heatmap highlights lower-performing models in red and higher-performing models in green.
As shown, eSpeak and HomoFast eSpeak are the fastest models, with the latter benefiting from a newly added feature that enables processing of larger text segments in a single run.


\section{Conclusion}
In this work, we tackled two persistent challenges in homograph disambiguation for low-resource languages: the high cost of dataset creation and the latency constraints of real-time G2P applications. We proposed a semi-automated pipeline for building homograph-rich datasets and introduced HomoRich, the first large-scale, openly licensed Persian homograph dataset. Using this resource, we achieved a 29.72\% improvement in homograph accuracy for a state-of-the-art neural G2P model.

To bridge the gap between accuracy and real-time performance, we further developed a lightweight, context-aware statistical method that enhances homograph handling with minimal computational overhead. Integrated into the widely used eSpeak engine, this method led to HomoFast eSpeak, a fast, homograph-aware G2P system that improves disambiguation accuracy by 30.66\% while retaining the responsiveness crucial for screen readers and other accessibility tools.

Our results highlight the potential of using high-quality offline datasets not only to train neural models, but also to enrich and modernize traditional rule-based systems. By releasing all resources under a CC0-1.0 license, we aim to foster further research and practical adoption in accessibility technologies for low-resource languages.

\section{Limitations}
Homograph disambiguation is not the only context-dependent challenge in Persian. Another notable challenge is the correct phonemization of the Ezafe, a linking phoneme that grammatically and semantically connects words. This is a major weakness in current rule-based systems.

Addressing such context-sensitive phenomena requires further research, particularly in designing fast yet linguistically aware rule-based methods. Tackling challenges like Ezafe handling could bring rule-based G2P models significantly closer to the naturalness of neural models—while maintaining the speed advantage crucial for real-world deployment.




\bibliography{custom}

\appendix

\section{Extended Related Work Review}
\label{appendix:extended-related-work}

This appendix provides a more detailed review of prior work on homograph disambiguation. We organize the discussion into two parts: first, we survey general approaches used across languages, including rule-based, statistical, neural, and hybrid methods. Then, we turn our focus to Persian-specific efforts, particularly those that involve the creation or use of datasets aimed at addressing the scarcity of resources for homograph disambiguation in low-resource settings.

\subsection{Homograph Disambiguation Approaches}

This subsection reviews the main approaches proposed for homograph disambiguation across languages. To provide a clear structure, we divide the methods into five categories: rule-based, neural, hybrid, LLM-based, and other approaches. This organization reflects both the chronological evolution and methodological diversity of the field.

\subsubsection{Rule-based Approaches}

\citet{silva2012rule} proposed a rule-based algorithm set as their core method for homograph disambiguation in Brazilian Portuguese text-to-speech systems. Their approach utilizes linguistic rules based on morphosyntactic and semantic analysis, employing information from the surrounding context, including part-of-speech, morphology, lemmas, and semantic relations from Wordnets, along with restrict lexical combinations. The authors tested their algorithms on existing text databases, namely a newspaper corpus (CETENFolha), the Holy Bible in BP, and Brazilian literature.

\citet{yarowsky1997homograph} developed a corpus-driven approach for English homograph disambiguation, utilizing a 400-million-word multi-domain dataset that included news articles, scientific texts, and literary works. Their method employed statistical decision lists that ranked contextual patterns (including adjacent words and part-of-speech tags) by their log-likelihood ratios to determine correct pronunciations, effectively addressing seven major categories of homographs through data-driven rules rather than neural networks. The work demonstrated how large-scale, diverse training data could be leveraged to resolve lexical ambiguities with high accuracy.

\citet{hearst1991noun} proposed a method for noun homograph disambiguation in English using a large unrestricted text corpus, the Academic American Encyclopedia, which contains approximately 8.6 million words. To address the lack of sense-annotated data, the author manually labeled a small set of training instances for each homograph—testing the method on five English nouns (e.g., bank, tank, bass)—and further improved performance through an unsupervised learning phase that incorporated high-confidence predictions without additional manual effort. The core method, called CatchWord, is rule-based and relies on shallow contextual cues such as syntactic patterns, orthographic features (e.g., capitalization), and lexical co-occurrence information extracted from local context windows. This approach avoids deep semantic resources or inference and demonstrates that coarse-grained disambiguation can be effectively achieved using lightweight, corpus-driven statistical techniques.

\subsubsection{Neural Approaches}

\citet{nicolis2021homograph} proposed a homograph disambiguation system for American English text-to-speech (TTS) applications, focusing primarily on neural methods rather than rule-based ones. They used a publicly available dataset comprising 138 homograph words, each with around 90 training and 10 test sentences, and addressed data imbalance by manually augmenting the training set for underrepresented homograph variants using an internal fiction-based corpus. This augmentation, which added about 10 examples per weak variant, led to a relative accuracy improvement of over 11\%, demonstrating the effectiveness of targeted data enrichment. Their method relies on contextual word embeddings (CWEs) extracted from pretrained BERT and ALBERT models, which are then fed into lightweight logistic regression classifiers trained separately for each homograph. This fully ML-based approach achieves state-of-the-art performance without the need for hand-crafted rules.

\citet{seale2021label} addressed the challenge of low-resource data in homograph disambiguation by exploring label imputation techniques. To mitigate this, the author generated four homograph disambiguation datasets and made them available for the research community. The author also used the Wikipedia Homograph Data (WHD) released by Gorman et al. (2018) to conduct the research. Their core method involved employing regularized, multinomial logistic regression and fine-tuning pre-trained ALBERT, BERT, and XLNet language models as token classifiers to improve model performance, particularly in classes with low prevalence samples.

\citet{ploujnikov2024towards} proposed SoundChoice, a sentence-level Grapheme-to-Phoneme (G2P) model aimed at improving homograph disambiguation in English. To address the challenge of context-aware phoneme prediction, they constructed the LibriG2P dataset, which integrates lexicon-based word pronunciations from CMUDict, phoneme alignments from LibriSpeech, and Wikipedia homograph data. This dataset includes approximately 10259 homograph-labeled samples, addressing inconsistencies between lexicon-based and audio-derived phoneme annotations. Their model employs a hybrid neural architecture, leveraging LSTMs, GRUs, and content-based attention, alongside CTC loss and curriculum learning—progressing from individual word training to sentence-level fine-tuning for enhanced contextual phoneme prediction. Additionally, BERT word embeddings are incorporated to inject semantic knowledge for better homograph resolution, achieving a phoneme error rate (PER) of 2.65\% and 94\% homograph classification accuracy. This work contributes to dataset development and model innovations in grapheme-to-phoneme conversion.

\citet{vrezavckova2024t5g2p}, \citet{vrezavckova2024homograph} introduced a grapheme-to-phoneme (G2P) conversion approach using a Text-to-Text Transfer Transformer (T5) model. To capture cross-word context and assimilation effects, their models for English and Czech were trained on proprietary datasets of several hundred thousand sentences provided by language experts, mitigating the need for explicit rule-based post-processing. The T5-based model achieved high conversion accuracy across the tested languages.

\citet{comini2025lightweight} present a neural-based lightweight front-end for on-device TTS in English, Polish, and Russian, using internal pronunciation dictionaries and the Kaggle text normalization dataset to address data limitations. Their dataset includes 53.1k, 42.4k, and 31.9k words for G2P and 6.4k, 6.5k, and 11.2k tokens for TN. They employ transformer-based and GRU-based models, leveraging knowledge distillation from pre-trained teacher models to train compact student models, optimizing for low latency and scalability in low-resource scenarios.

\citet{gao2024unsupervised} tackle speech processing for low-resource languages using neural methods, particularly self-supervised learning (SSL) with models like wav2vec2 and HuBERT. They use existing speech datasets (e.g., LibriSpeech, VoxPopuli, CommonVoice) and enhance SSL pretraining with synthetic speech generated by diffusion models to address data scarcity. Their approach improves multilingual and zero-shot phonetic recognition without requiring labeled data.

\citet{nanni2023disambiguating} investigated homographic heterophone disambiguation in Italian Text-To-Speech (TTS) systems using the SoundChoice model, which includes an RNN (LSTM + GRU) and a transformer version. Given the scarcity of Italian homograph datasets, the study generated 9,916 sentences with ChatGPT, supplementing a 1,700-sentence corpus dataset. The ChatGPT-generated data was created through iterative prompting, where sentences were crafted to include homographs in varying syntactic contexts. These sentences were manually validated for linguistic accuracy and context relevance before phonetic transcription using a ReadSpeaker transcription tool, which had a 59.56\% accuracy in homograph resolution. The model integrates semantic disambiguation via BERT embeddings and a weighted homograph loss, enabling sentence-level pronunciation prediction. Evaluation showed the transformer model outperformed the RNN, highlighting the feasibility of neural methods for Italian homograph disambiguation.

\subsubsection{Hybrid Approaches}

\citet{gorman2018improving} addressed homograph disambiguation for English TTS by creating a labeled dataset of 163 homographs (including morphosyntactic, lexical, and mixed types), with ~100 sentences per homograph sampled from Wikipedia and annotated via crowdsourcing. To mitigate data scarcity, they employed rigorous adjudication for label disagreements and released the dataset publicly. Their hybrid system combined rule-based heuristics (e.g., context-triggered pronunciation rules, POS tags) with supervised ML (per-homograph maxent classifiers using word-context, POS, and capitalization features), showing that hybridization outperformed either approach alone.

\citet{karamihaylova2023neural} developed a hybrid grapheme-to-phoneme (G2P) system for Bulgarian, combining rule-based finite-state transducers (FSTs) for consonant mapping and vowel reduction rules with an LSTM-based seq2seq model for stress prediction. To address inconsistencies in publicly available data, they scraped and filtered ~38,000 word-pronunciation pairs from Bulgarian Wiktionary using WikiPron, then standardized consonant transcriptions while preserving vowel variations to study stress-induced reduction. The dataset included homographs, where stress position disambiguates meaning. Their hybrid approach achieved performance comparable to pure neural methods, demonstrating the viability of curated rule-neural integration for medium-resource languages.

\subsubsection{LLM-based Approaches}

\citet{suvarna2024phonologybench} introduced PhonologyBench, evaluating Large Language Models (LLMs) on English phonological tasks, including homographs. Their dataset includes 3,000 words for grapheme-to-phoneme conversion, sourced from SIGMORPHON 2021, ensuring phonemic transcriptions. They tested GPT-4, Claude-3-Sonnet, and LLaMA-2-13B, using a zero-shot neural approach, showing that LLMs struggle with homograph pronunciation. Their findings highlight the need for phonology-aware datasets to improve text-based pronunciation models.

\citet{han2024improving} explored the use of Large Language Models (LLMs) for grapheme-to-phoneme conversion, focusing on leveraging the in-context knowledge retrieval capabilities of GPT-4 to disambiguate homographs. To facilitate this, the authors constructed a dictionary by combining the Librig2p training dataset and the CMU dictionary. For homograph words, they used GPT-4 to generate cases automatically. Each homograph contains multiple cases and was later manually refined. The core of their method involves prompting GPT-4 to analyze the input sentence, identify the most relevant meaning and part-of-speech for the target word, and then retrieve the corresponding phoneme pronunciation from the constructed dictionary.

\citet{qharabagh2025llm} In a previous study, we proposed an LLM-powered approach to Grapheme-to-Phoneme (G2P) conversion in Persian, addressing challenges posed by polyphone words and context-sensitive phonemes. To improve phonetic accuracy and benchmark sentence-level G2P performance, we introduced two datasets: Kaamel-Dict, a unified phonetic dictionary with 120,000+ entries, and Sentence-Bench, a sentence-level dataset containing 400 annotated sentences, including 100 polyphone words used in various contexts. Our method leverages large language models (LLMs) without additional training, applying advanced prompting and post-processing techniques to enhance phonetic predictions. Our benchmarking results demonstrate that LLMs can outperform traditional models, highlighting the potential of LLMs in low-resource G2P tasks.

\subsubsection{Other Approaches}

\citet{tesprasit2003context} addressed the challenges posed by word boundary and homograph ambiguity in Thai Text-to-Speech, noting the absence of word delimiters in the language. To conduct their research, they created their own 25K-word corpus where sentences were manually segmented, and part-of-speech tags and pronunciations were manually annotated by linguists. Their core method is a unified machine learning framework based on the Winnow algorithm, a statistical technique that learns to combine local and long-distance contextual features like context words and collocations to disambiguate word pronunciations without relying on predefined rules or standard neural network architectures.

\citet{alqahtani2019homograph} addressed homograph disambiguation in Arabic by proposing unsupervised, data-driven methods to selectively restore diacritics, balancing lexical disambiguation and sparsity. They leveraged existing corpora (~50M tokens, including Gigaword and Arabic Treebank) without new data collection, using the MADAMIRA tool for automated diacritization and morphological analysis. Their approach identified 33.8\% of words as homographs (e.g., 168K ambiguous types) by clustering diacritized variants (Brown, K-means) and analyzing translation divergences in parallel text. Unlike rule-based or neural methods, their work focused on distributional similarity and morphological variants to guide selective diacritization, demonstrating improved performance in downstream tasks like machine translation and POS tagging.

\citet{hajj2022comparing} addressed the challenge of disambiguating French heterophonic homographs for TTS systems by creating a custom dataset. They collected 8137 sentences from the web, ensuring a balanced representation of 34 pairs of prototypical homographs, with roughly one hundred instances per pair. To enhance disambiguation, they employed Linear Discriminant Analysis (LDA) classifiers, utilizing contextual word embeddings as input features, and experimented with the FlauBERT transformer for POS tagging.

\subsection{Persian Homograph Disambiguation and Dataset Development}
\label{sec:appendix-datasets-review}

Several recent works have introduced or curated datasets specifically for Persian homograph disambiguation and word sense disambiguation (WSD). Notably, \citet{moghadaszadeh2024avashog2p} presented a dataset collected through a cluster-based sampling strategy to mitigate phoneme imbalance. Another valuable dataset is by \citet{ghayoomi2019identifying}, who developed a manually annotated gold standard for 20 Persian ambiguous words, each with 100 sentences, totaling 2000 sentences. These sentences were extracted from the Persian Language Database and annotated according to SemEval2010 guidelines. Similarly, \citet{rahmati2024ge2pe} generated over 5 million sentence-phoneme pairs, including manually and automatically labeled data which was a valuable source for general G2P task not homograph challenge.

Other works focused on smaller, curated datasets. \citet{ayyoubzadeh2024persian} created a dataset containing 63 homograph words, with sentence-level phonetic annotations developed through careful selection. \citet{mahmoodvand2017semi} extracted 5368 documents/sentences using a web crawler for three Persian homographs ("Shir", "Rast", "Tar") from Iranian news agency websites, partially labeled (2133 documents). \citet{riahi2012semi} used the Hamshahri corpus and manually tagged instances of two homographs, with training sizes ranging from 10 to 1500 words for their tri-training framework. Additionally, \citet{nanni2023disambiguating} created an Italian homograph dataset, including 9,916 ChatGPT-generated sentences supplemented with 1,700 corpus examples.

The following paragraphs provide a more detailed examination of each study.

\citet{riahi2012semi} addressed the challenge of limited manually tagged data for Persian Word Sense Disambiguation (WSD) by proposing a semi-supervised method. To conduct their experiments, they utilized the raw Hamshahri corpus and created their own tagged data by manually annotating instances of two Persian homographs. Their core method employs a statistical approach based on tri-training with decision lists. The decision lists classify homographs by analyzing the distribution of collocations (surrounding words), and the tri-training framework iteratively leverages a small tagged corpus and a larger untagged corpus to improve disambiguation accuracy.

\citet{moghadaszadeh2024avashog2p} introduced AvashoG2P, a multi-module system for Persian grapheme-to-phoneme (G2P) conversion that primarily employs neural network-based approaches. For out-of-vocabulary word prediction, their core method utilizes a sequence-to-sequence model with a GRU-based recurrent unit and an attention mechanism. Addressing the lack of labeled data for homograph disambiguation in Persian, the authors collected and labeled their own homograph data. To mitigate the challenge of data imbalance in their collected homograph data, they first clustered the data for each homograph before labeling a selection of samples from each cluster. Their homograph disambiguation module leverages a classification approach that uses a single model for all 54 supported Persian homographs, with experiments highlighting the superior performance of transformer-based models like XLMRoberta.

\citet{ghayoomi2019identifying} proposed an unsupervised neural method for Persian word sense induction using word embeddings and hierarchical clustering. They trained embeddings on a combined corpus (~529M words) and evaluated on a manually annotated dataset of 20 ambiguous words (100 sentences each). Their approach leveraged context windows (8 surrounding words) and sentence-level embeddings, clustering them without predefined rules.

\citet{ayyoubzadeh2024persian} introduced a novel dataset for Persian homograph disambiguation, addressing the challenges posed by words with identical spellings but different meanings in Persian[1]. Their dataset includes diverse sentences containing homographs, which are carefully annotated to facilitate detailed analysis and model training. The authors trained both lightweight machine learning and deep learning models, leveraging embeddings and cosine similarity to disambiguate homographs and evaluated model performance using accuracy, recall, and F1 score.

\citet{mahmoodvand2017semi} addressed the challenge of limited labeled data for Persian word sense disambiguation by implementing a semi-supervised machine learning approach. They created their own corpus by developing a crawler to extract sentences containing target ambiguous words from news agency websites, building a dataset specifically designed for WSD tasks. Their method leverages a small set of labeled seed data combined with a larger volume of unlabeled data in a collaborative learning framework, focusing on defined features of target words to disambiguate their meanings. The researchers evaluated their approach on three Persian homograph words ("Shir," "Rast," and "Tar"), achieving impressive results with 88\% recall, 95\% precision, and 93\% accuracy across 5,368 documents, demonstrating the effectiveness of their semi-supervised approach for Persian language processing despite the inherent challenges of Persian's rich metaphorical nature and complex writing style.

\citet{mahmoodvand2015persian} presented a method for building a Persian word sense disambiguation (WSD) dataset by employing a web crawler to gather documents containing specific ambiguous words. Addressing the lack of suitable WSD corpora for Persian, their approach focuses on extracting relevant phrases for ambiguous words from web data to create a dataset that can be used in WSD tasks. The authors used three prevalent Persian ambiguous words to extract appropriate phrases. This research provides a foundation for supervised WSD methods in Persian by offering a means to generate training data where it was previously scarce.

\citet{rahmati2024ge2pe} proposed GE2PE, a Persian end-to-end grapheme-to-phoneme conversion model that addresses the challenges of Persian homographs and missing short vowels by leveraging sentence-level context. To support this, they created two large datasets comprising over five million sentences with corresponding phoneme sequences, including both manually labeled and machine-generated data, and designed evaluation sets specifically for tasks like Kasre-Ezafe detection and homograph disambiguation. Their core approach is a T5 model trained in a two-step process, building on advances in transformer architectures shown to be effective for G2P tasks. This work stands out for its extensive data creation tailored to Persian linguistic challenges and its end-to-end neural modeling strategy.

\citet{rezaei2022multi} proposed a multi-module G2P system for Persian that addresses the challenges of homographs, OOV words, and ezafe constructions. To handle homographs, they extracted a homograph dictionary from the Ariana lexicon. Their core method involves a combination of GRU and Transformer architectures within separate modules to handle different aspects of G2P conversion. The system operates at the sequence level, capturing cross-word relations crucial for homograph disambiguation and ezafe recognition.

\citet{alayiaboozar2019word} proposed a rule-based approach for disambiguating Persian noun and adjective homographs ending in (/i/), leveraging context-sensitive syntactic rules (e.g., preposition + quantifier patterns) derived from three existing corpora: the Peykare corpus, Farsi Linguistic Database, and Persian Dependency Treebank. They extracted 36 rules based on 10-word contextual windows, achieving high accuracy (e.g., 94\% for some rules), but did not create new labeled data. Their method focused on morphological and syntactic patterns (e.g., adjacent POS tags) to resolve ambiguity in a language with prevalent homography due to orthographic constraints.

\section{Phoneme Representaion Mapping}
\label{sec:mapping}

\begin{figure*}[t]
    \centering
    \begin{subfigure}[b]{0.48\textwidth}
        \centering
        \includegraphics[width=\textwidth]{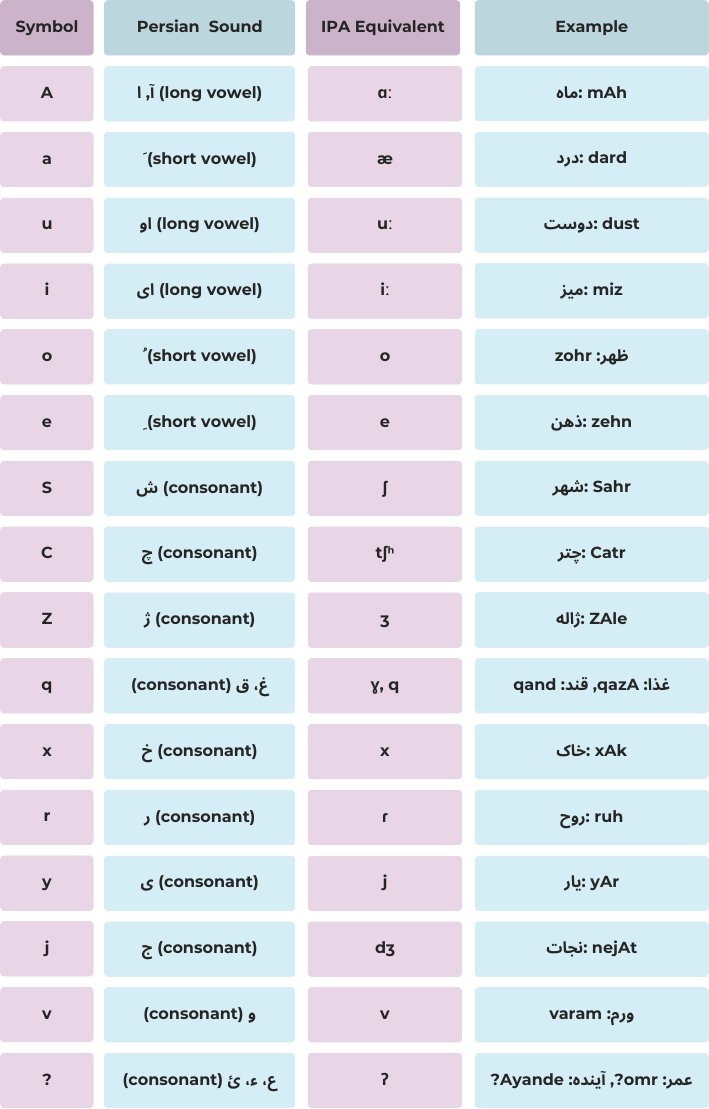}
        \caption{Repr. 1 (used in this work and related studies)}
        \label{fig:ours-phoneme-rep}
    \end{subfigure}
    \hfill
    \begin{subfigure}[b]{0.48\textwidth}
        \centering
        \includegraphics[width=\textwidth]{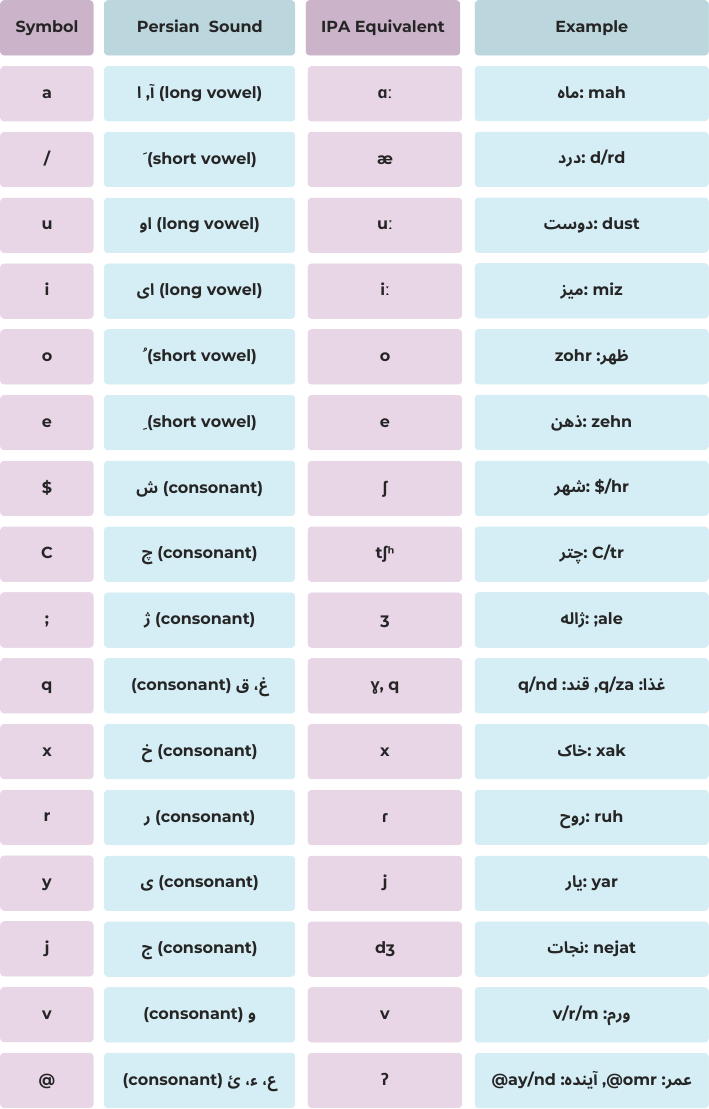}
        \caption{Repr 2 (used in other prior literature)}
        \label{fig:alt-phoneme-rep}
    \end{subfigure}
    \caption{Comparison of two commonly used phoneme representations for Persian sounds.}
    \label{fig:phoneme-rep}
\end{figure*}

There are two common representations for Persian phonemics. The first representation is the one used in many of the G2P glossaries, including KaamelDict \cite{fetrat_kaameldict, tihu-dict, IPA-Translator, wiktionary, zaya, jame, mohammadhasan, PasaOpasen, azamrabiee, zhu2022charsiu-g2p} and benchmarks like SentenceBench \cite{fetrat2024sentencebench}.  
The second representation is used in one of the state-of-the-art G2P models for Persian, namely GE2PE, which is fine-tuned and enhanced in this work. Our HomoRich dataset includes the sentence phoneme sequences in both of these formats for compatibility. Figure~\ref{fig:phoneme-rep} shows these two representations.

A key challenge in mapping the Ezafe phoneme between these representations was its inconsistent annotation. The Ezafe is a short vowel /e/ used to indicate possession, relation, or description in Persian noun phrases. For instance, in the sentences "This is Ziba’s flower" (/in gol-e zibA ast/) and "This flower is beautiful" (/in gol zibA ast/), the Ezafe appears as a linking /e/ sound, but its presence or absence can alter the meaning of the sentence.  
In the GE2PE representation, the Ezafe is denoted by an additional `1` symbol after the `/e/` phoneme, while in our dataset, the `/e/` phoneme alone may indicate either a regular vowel or an Ezafe.  

To resolve this ambiguity, we employed a POS tagger \cite{hazm} with 99.249\% accuracy to identify Ezafe constructions based on the grapheme sequence. For each Ezafe occurrence, we retrieved its phonemic form from the KaamelDict \cite{fetrat_kaameldict} glossary and searched for the corresponding `/e/` phoneme in the phoneme sequence. A `1` symbol was then appended to the `/e/` to maintain consistency with the GE2PE representation.

\section{Additional Figures}

\begin{figure*}[t]
\centering
\includegraphics[width=\textwidth]{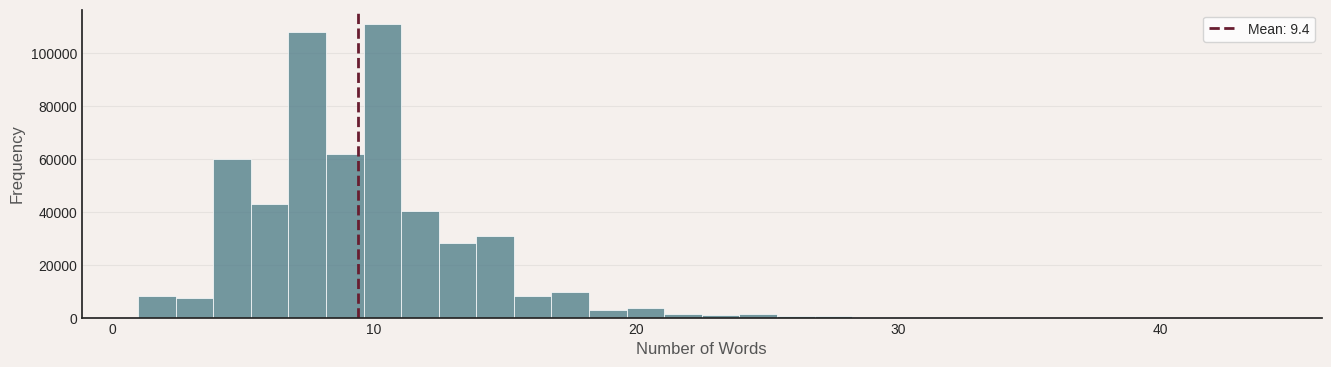}
\caption{Distribution of sentence word counts.}
\label{fig:word-count-hist}
\end{figure*}

\paragraph{Dataset Sentence Length Distribution}
A well-designed dataset for a G2P model should include sentences of varying lengths to ensure the model can accurately transcribe both short and long utterances. Sentence length is also an indicator of linguistic diversity and complexity. Figure~\ref{fig:word-count-hist} illustrates the distribution of sentence lengths in the Homorich dataset.

\begin{figure*}[t]
\centering
\begin{subfigure}[b]{0.3\textwidth}
\includegraphics[width=\textwidth]{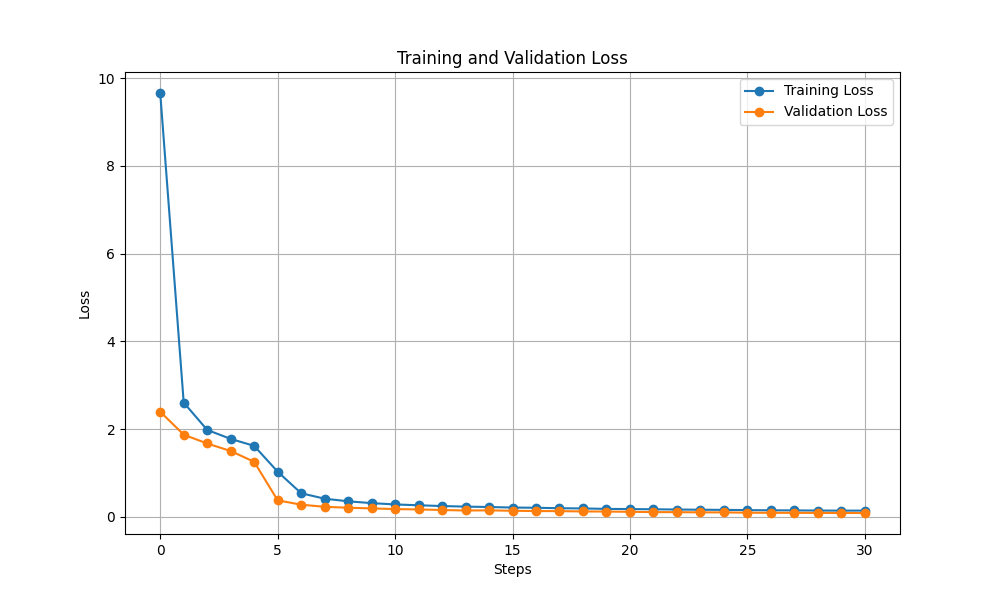}
\caption{Phase 1 (5 epochs)}
\label{fig:phase1}
\end{subfigure}
\hfill
\begin{subfigure}[b]{0.3\textwidth}
\includegraphics[width=\textwidth]{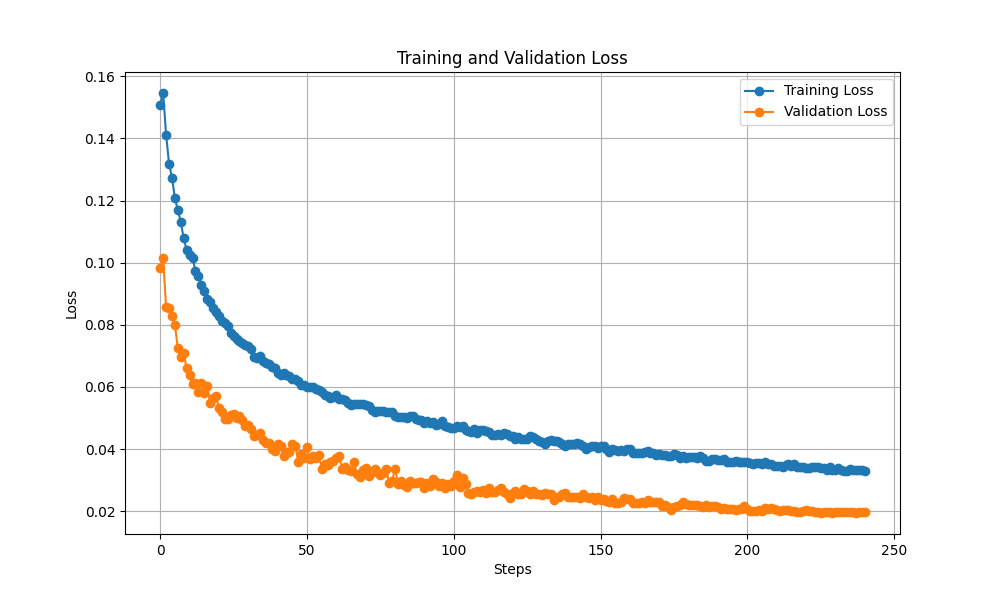}
\caption{Phase 2 (20 epochs)}
\label{fig:phase2}
\end{subfigure}
\hfill
\begin{subfigure}[b]{0.3\textwidth}
\includegraphics[width=\textwidth]{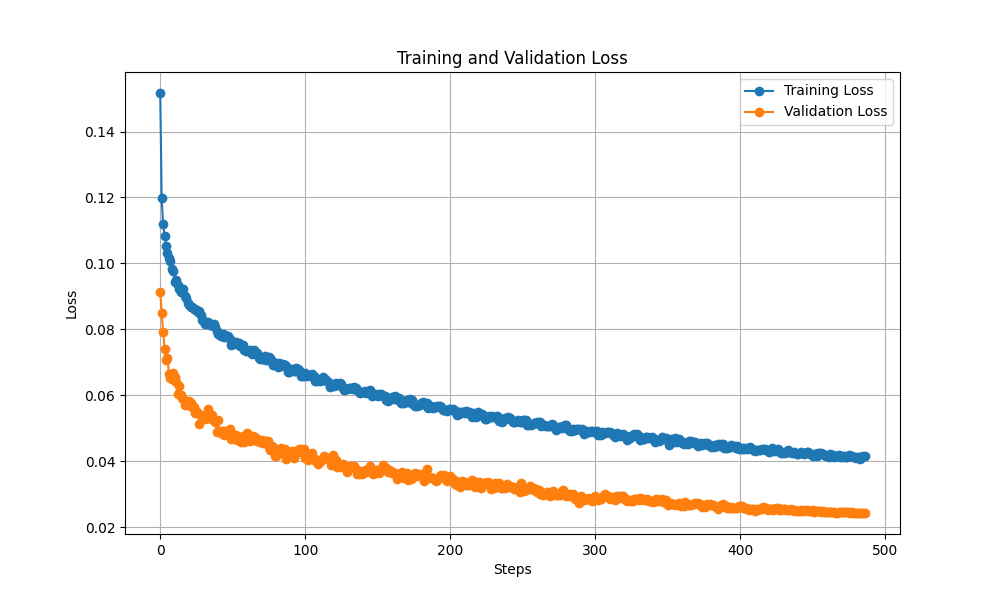}
\caption{Phase 3 (50 epochs)}
\label{fig:phase3}
\end{subfigure}
\caption{Learning curves across fine-tuning phases of T5.}
\label{fig:t5-learning-curves}
\end{figure*}

\paragraph{Learning Curves for T5 Training}
Figure~\ref{fig:t5-learning-curves} shows the training dynamics across all phases when fine-tuning T5, with identical hyperparameters as described in Section~\ref{sec:homo-ge2pe}.

\section{Statistical Analysis of Experimental Results}
\label{sec:error-bar}

\begin{figure*}[t]
    \centering
    \includegraphics[width=\textwidth]{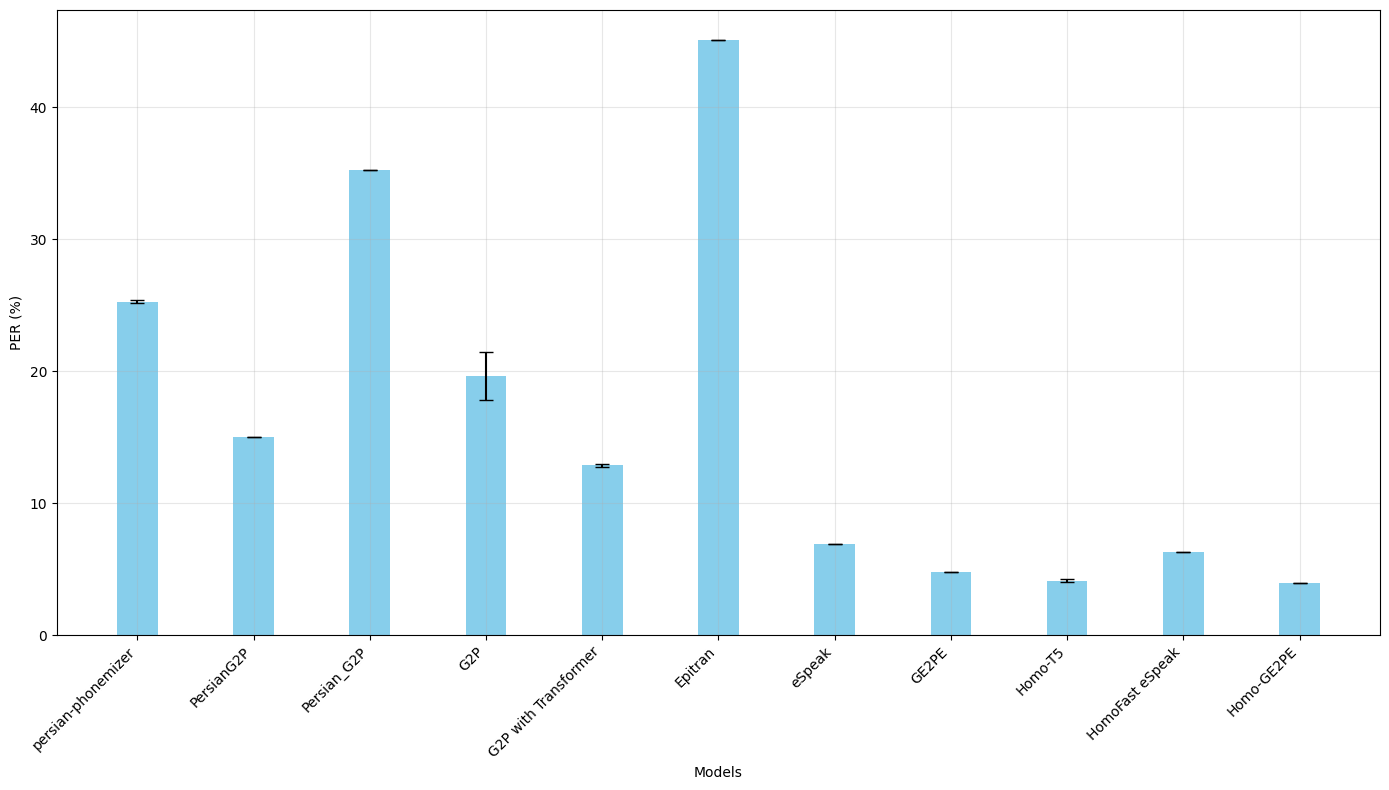}
    \caption{Phoneme Error Rate (PER) of previous and proposed G2P tools/models with error bars indicating standard deviations across five runs.}
    \label{fig:per}
\end{figure*}

\begin{figure*}[t]
    \centering
    \includegraphics[width=\textwidth]{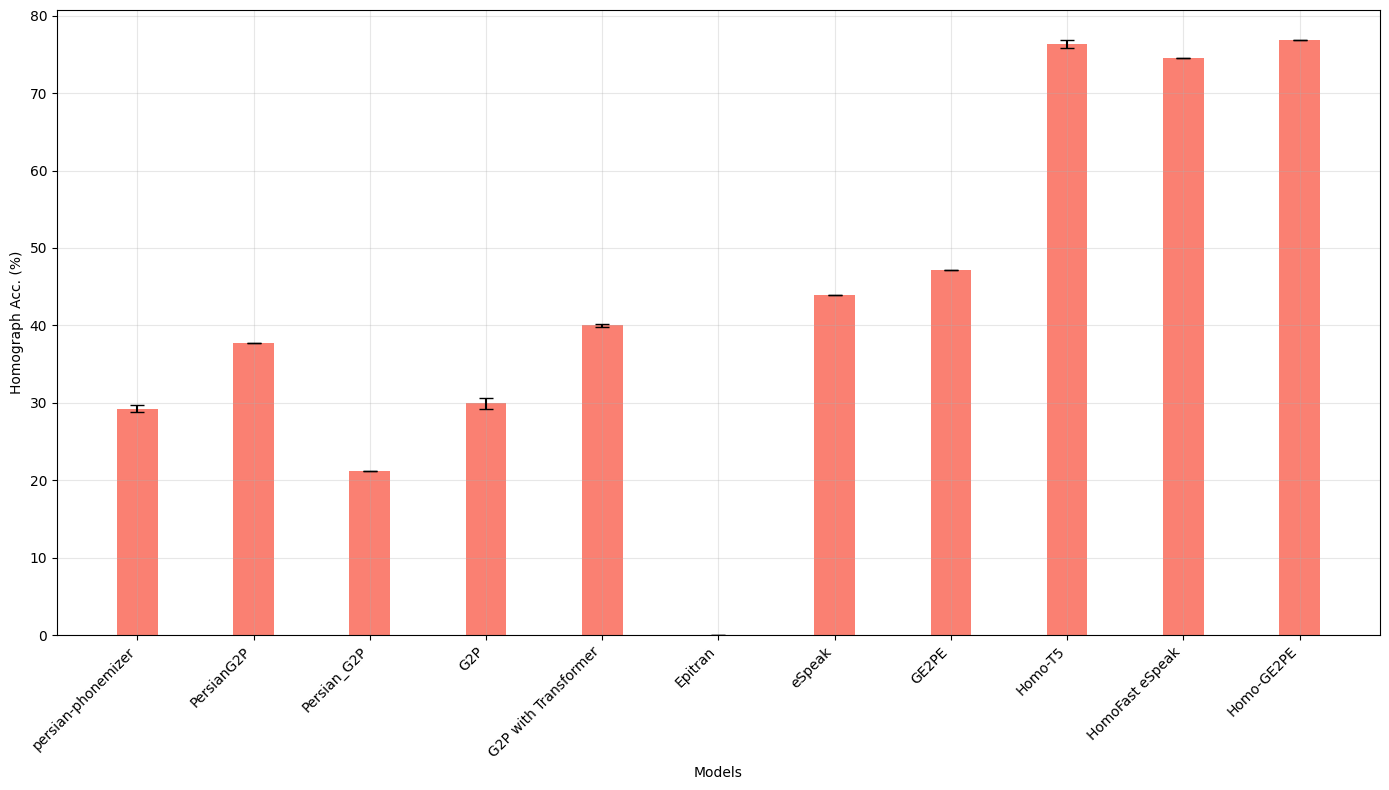}
    \caption{Homograph Accuracy of previous and proposed G2P tools/models with error bars indicating standard deviations across five runs.}
    \label{fig:homograph}
\end{figure*}

\begin{figure*}[t]
    \centering
    \includegraphics[width=\textwidth]{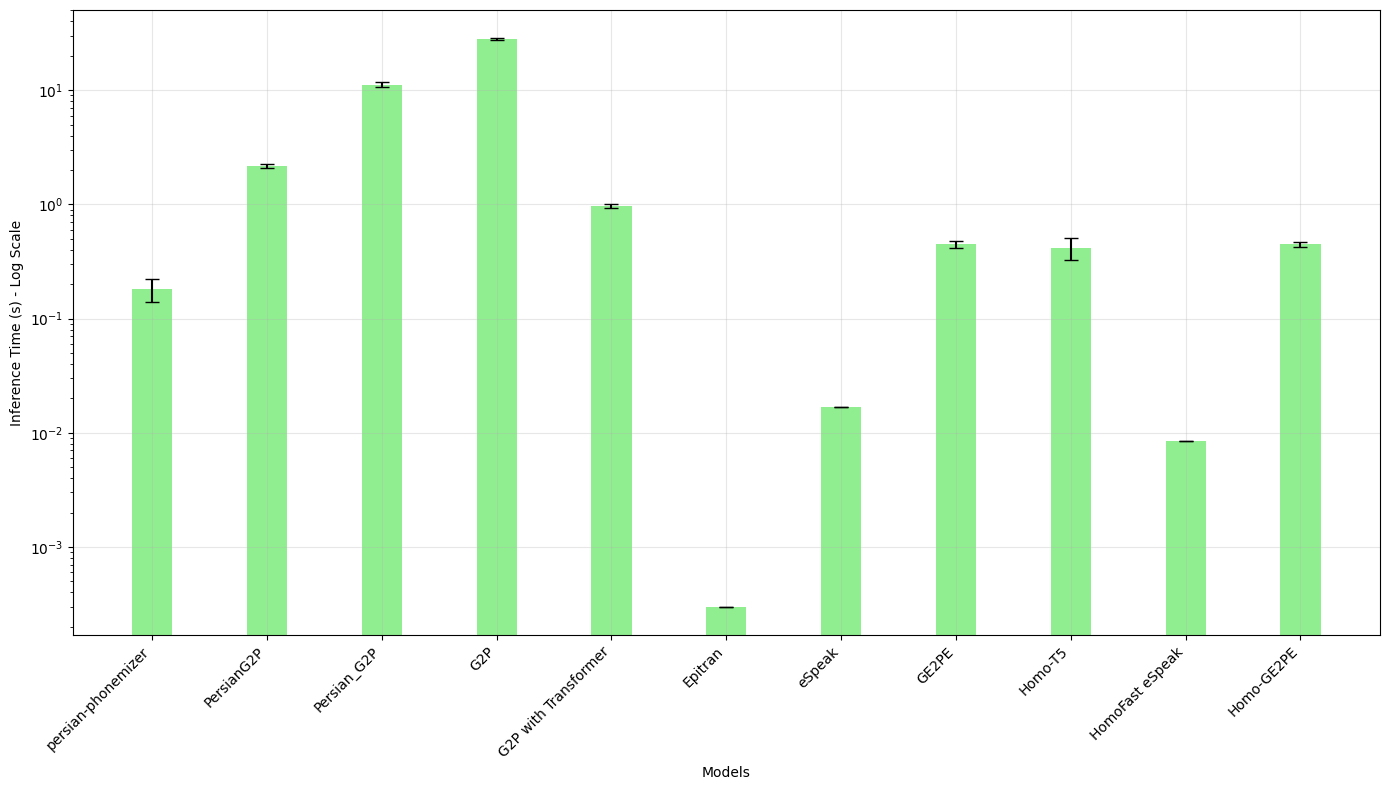}
    \caption{Inference Time (s) of previous and proposed G2P tools/models plotted on a logarithmic scale with error bars indicating standard deviations across five runs.}
    \label{fig:time}
\end{figure*}

To provide a comprehensive view of the variability in the reported metrics, we present error bar plots for the Phoneme Error Rate (PER), Homograph Accuracy, and Inference Time across the evaluated G2P tools and proposed models. Figures~\ref{fig:per}, \ref{fig:homograph}, and \ref{fig:time} illustrate these metrics, with error bars representing standard deviations across five runs. The inference time plot is rendered on a logarithmic scale to highlight differences across models with varying computational requirements.

\section{Details of Human Subject Participation in Data Collection}
\label{sec:human-subject}

As part of this study, we engaged approximately 200 human participants to contribute to the human-generated portion of our homograph sentence corpus. Specifically, we curated a list of 285 Persian homograph words, each with multiple valid pronunciations. These were organized into several Google Sheets, where each sheet listed a subset of homograph words along with their pronunciations, followed by five empty rows designated for sentence creation for each alternative.

Each homograph word appeared in multiple sheets to ensure that it was annotated by different individuals, and each participant received a subset of words—thus distributing the workload and encouraging diversity in linguistic expression. The instructions, originally provided in Persian, asked participants to compose Persian sentences that naturally incorporate the target homograph with the specified pronunciation. A translated excerpt of the instruction reads:

\begin{quote}
"Please write five different Persian sentences using the given word with the pronunciation indicated below it. Try to make the sentences as natural and diverse as possible. Avoid repeating sentence structures or vocabulary."
\end{quote}

Participants were explicitly encouraged to avoid sentence repetition, maintain lexical diversity, and write fluent, meaningful examples.

Each participant completed multiple such entries, and collectively, this process yielded a total of 69,560 high-quality, human-written sentences. The sentences form a valuable component of our dataset for disambiguating homograph pronunciation in context.

\section{Broader Impact}
\label{sec:broader-impact}

The ultimate goal of our work is to improve the quality of fast, rule-based G2P models—and neural G2P systems in general—so they can be effectively integrated into low-latency text-to-speech (TTS) pipelines, particularly for screen readers and other real-time accessibility tools. By enhancing homograph disambiguation and overall phonetic accuracy, we enable more natural and reliable speech synthesis, which is critical for users who rely on assistive technologies.

A key practical outcome of our research is the development of HomoFast eSpeak, an enhanced version of the widely used open-source eSpeak NG speech synthesizer. Our experiments show that HomoFast eSpeak achieves a 30.66\% improvement in homograph disambiguation accuracy while maintaining the low-latency performance critical for real-time applications. This advancement has the potential to elevate the intelligibility and naturalness of synthesized speech in screen readers used by the blind community in Iran.

Beyond immediate applications, we hope this work encourages further development of open, high-quality, and performant TTS systems for low-resource languages. By releasing our dataset (HomoRich), models (Homo-GE2PE), and enhancements to eSpeak under permissive licenses, we lower barriers for researchers and developers working on accessibility-focused speech technologies. Our contributions demonstrate that even simple, data-informed statistical methods can significantly improve rule-based systems—making high-quality G2P more scalable and sustainable for languages with limited resources.

\section{Disclosure of LLM usage}
We used large language models (LLMs) for language refinement, including grammar correction, paragraph rephrasing, and other minor edits, based on drafts written by the authors. In the related work section, LLMs assisted in summarizing prior works after the authors had identified, read summaries of, and grouped the relevant literature; this use was limited to generating low-novelty text describing pre-existing methods and data. The generated text was subsequently reviewed for accuracy. Additionally, LLMs were used for fine-grained coding tasks such as generating individual functions or single-purpose scripts, which were then validated and integrated by the authors.

\section{Data Sheet}
In the rest of this document, we present the datasheet for the HomoRich dataset, adhering to the guidelines outlined by \citet{gebru2021datasheets}.

\subsection{Motivation}

The questions in this section are primarily intended to encourage
dataset creators to clearly articulate their reasons for creating the
dataset and to promote transparency about funding interests.
The latter may be particularly relevant for datasets created for
research purposes.\\

\paragraph{For what purpose was the dataset created?} Was there a specific task in mind? Was there a specific gap that needed to be filled? Please provide a description.

\paragraph{ANS:} The dataset was created to address the scarcity of open-source datasets and models for grapheme-to-phoneme (G2P) conversion, with a focus on homograph disambiguation in Persian. These resources aim to support the development of open text-to-speech (TTS) and screen reader tools, enhancing accessibility for Persian-speaking communities, including individuals with visual impairments.

\paragraph{Who created the dataset (e.g., which team, research group) and on behalf of which entity (e.g., company, institution, organization)?}

\paragraph{ANS:} The dataset was created by the speech processing team of the Data Science and Machine Learning (DML) Laboratory at Sharif University of Technology.

\paragraph{Who funded the creation of the dataset?} If there is an associated grant, please provide the name of the grantor and the grant name and number.

\paragraph{ANS:} The dataset creation received no external funding and is provided free of charge.

\paragraph{Any other comments?}

\paragraph{ANS:} No.

\subsection{Composition}

Most of the questions in this section are intended to
provide dataset consumers with the information they need to make
informed decisions about using the dataset for their chosen
tasks. Some of the questions are designed to elicit information
about compliance with the EU's General Data Protection Regulation
(GDPR) or comparable regulations in other jurisdictions.

\paragraph{What do the instances that comprise the dataset
    represent (e.g., documents, photos, people, countries)?} Are there
  multiple types of instances (e.g., movies, users, and ratings;
  people and interactions between them; nodes and edges)? Please
  provide a description.

\paragraph{ANS:} The dataset consists of Persian sentences (text) paired with their corresponding phoneme sequences in two formats (text). A subset of the dataset includes carefully curated Persian sentences containing homograph words, where each homograph and its pronunciation are explicitly annotated (text). All samples include metadata indicating their source (human, GPT-4o, CommonVoice, ManaTTS, or GPTInformal) and a unique identifier within each source category.

\paragraph{How many instances are there in total (of each type, if appropriate)?}

\paragraph{ANS:} The dataset contains 528,891 Persian sentences in total, with 327,475 specifically curated for homograph disambiguation.

\paragraph{Does the dataset contain all possible instances or is it
    a sample (not necessarily random) of instances from a larger set?}
  If the dataset is a sample, then what is the larger set? Is the
  sample representative of the larger set (e.g., geographic coverage)?
  If so, please describe how this representativeness was
  validated/verified. If it is not representative of the larger set,
  please describe why not (e.g., to cover a more diverse range of
  instances, because instances were withheld or unavailable).

\paragraph{ANS:} The dataset incorporates: (1) complete samples from ManaTTS and GPTInformal (covering all available data at study time), and (2) a non-random subset of CommonVoice selected by availability (prioritizing validated samples while respecting original data ordering). GPT-4o generations and human annotations were collected specifically for this study.

\paragraph{What data does each instance consist of?} ``Raw'' data
  (e.g., unprocessed text or images) or features? In either case,
  please provide a description.

\paragraph{ANS:} Each instance contains processed Persian text along with its corresponding phoneme sequence represented in two formats: a primary phonemic transcription and an alternative standardized representation mapped for compatibility. For instances containing homographs, the data additionally includes the identified homograph word and its correct pronunciation in both representation formats.

\paragraph{Is there a label or target associated with each
    instance?} If so, please provide a description.

\paragraph{ANS:} Yes, each instance serves multiple labeling purposes. The complete phoneme sequence of the sentence acts as the primary label. For homograph-containing instances, additional labels include the specific homograph word and its contextually correct pronunciation, enabling the dataset to support both general grapheme-to-phoneme conversion and specialized homograph disambiguation tasks.

\paragraph{Is any information missing from individual instances?}
  If so, please provide a description, explaining why this information
  is missing (e.g., because it was unavailable). This does not include
  intentionally removed information, but might include, e.g., redacted
  text.

\paragraph{ANS:} Due to our semi-automated data creation pipeline, sentences containing multiple homograph words only have the target homograph (the focus of that particular instance) annotated.

\paragraph{Are relationships between individual instances made
    explicit (e.g., users' movie ratings, social network links)?} If
  so, please describe how these relationships are made explicit.

\paragraph{ANS:} No.

\paragraph{Are there recommended data splits (e.g., training,
    development/validation, testing)?} If so, please provide a
  description of these splits, explaining the rationale behind them.

\paragraph{ANS:} The dataset does not come with predefined splits. We recommend using the entire dataset for training while evaluating performance on the dedicated SentenceBench test set, following the methodology established in our work. This approach ensures consistent benchmarking across studies.

\paragraph{Are there any errors, sources of noise, or redundancies
    in the dataset?} If so, please provide a description.

\paragraph{ANS:} As detailed in the data creation process, some sentences were generated by GPT-4o with prompts targeting specific homograph pronunciations. While we implemented techniques to prevent these issues, the approach carries inherent limitations including potential hallucinated sentences and occasional incorrect homograph usage. Additionally, phonemization was performed using the LLM-based method from prior work, which achieves a phoneme error rate of 6.43\% and homograph accuracy of 64\%, representing another source of potential noise in the phonetic transcriptions.

\paragraph{Is the dataset self-contained, or does it link to or
    otherwise rely on external resources (e.g., websites, tweets,
    other datasets)?} If it links to or relies on external resources,
    a) are there guarantees that they will exist, and remain constant,
    over time; b) are there official archival versions of the complete
    dataset (i.e., including the external resources as they existed at
    the time the dataset was created); c) are there any restrictions
    (e.g., licenses, fees) associated with any of the external
    resources that might apply to a dataset consumer? Please provide
    descriptions of all external resources and any restrictions
    associated with them, as well as links or other access points, as
    appropriate.

\paragraph{ANS:} The dataset is self-contained and doesn't rely on external resources.

\paragraph{Does the dataset contain data that might be considered
    confidential (e.g., data that is protected by legal privilege or
    by doctor--patient confidentiality, data that includes the content
    of individuals' non-public communications)?} If so, please provide
    a description.

\paragraph{ANS:} The dataset contains no confidential or personal information. All data originates from three sources: (1) established public datasets (CommonVoice, ManaTTS, and GPTInformal), (2) GPT-4o generated content, and (3) contributions from voluntary human participants who provided non-sensitive example sentences.

\paragraph{Does the dataset contain data that, if viewed directly,
    might be offensive, insulting, threatening, or might otherwise
    cause anxiety?} If so, please describe why.

\paragraph{ANS:} The dataset is derived from well-known public datasets, the safeguarded GPT-4o model, and voluntary human subjects in an academic environment who were specifically asked to generate example sentences. Given these controlled sources and collection methods, we believe it is unlikely to contain offensive or harmful content. However, as with any language dataset, we recommend users review the content for their specific application needs.

\paragraph{Does the dataset identify any subpopulations (e.g., by
    age, gender)?} If so, please describe how these subpopulations are
  identified and provide a description of their respective
  distributions within the dataset.

\paragraph{ANS:} The dataset does not identify any subpopulations.

\paragraph{Is it possible to identify individuals (i.e., one or
    more natural persons), either directly or indirectly (i.e., in
    combination with other data) from the dataset?} If so, please
    describe how.

\paragraph{ANS:} We believe identification is not possible, as the data consists of voluntarily provided sample sentences generated for specific words.

\paragraph{Does the dataset contain data that might be considered
    sensitive in any way (e.g., data that reveals race or ethnic
    origins, sexual orientations, religious beliefs, political
    opinions or union memberships, or locations; financial or health
    data; biometric or genetic data; forms of government
    identification, such as social security numbers; criminal
    history)? If so, please provide a description.}

\paragraph{ANS:} The dataset consists of linguistic examples derived from established public datasets, the safeguarded GPT-4o model, and voluntary contributions from participants in an academic setting. Given these controlled sources and collection methods focused solely on language patterns, we believe it is unlikely to contain sensitive information. However, as with any textual dataset, we recommend users assess the content for their specific requirements.

\paragraph{Any other comments?}

\paragraph{ANS:} No.

\subsection{Collection Process}

In addition to the goals outlined in the previous section, the
questions in this section are designed to elicit information that may
help researchers and practitioners to create alternative datasets with
similar characteristics.\\

\paragraph{How was the data associated with each instance
    acquired?} Was the data directly observable (e.g., raw text, movie
  ratings), reported by subjects (e.g., survey responses), or
  indirectly inferred/derived from other data (e.g., part-of-speech
  tags, model-based guesses for age or language)? If the data was reported
  by subjects or indirectly inferred/derived from other data, was the
  data validated/verified? If so, please describe how.

\paragraph{ANS:} The data combines three acquisition methods: (1) directly observable text from public datasets (CommonVoice, ManaTTS, GPTInformal), (2) GPT-4o-generated sentences with targeted homograph usage (indirectly derived through prompting), and (3) human-authored sentences voluntarily contributed in an academic setting. No specific validation was performed on the LLM-generated or human-provided data beyond the collection methods described in the paper.

\paragraph{What mechanisms or procedures were used to collect the
    data (e.g., hardware apparatuses or sensors, manual human
    curation, software programs, software APIs)?} How were these
    mechanisms or procedures validated?

\paragraph{ANS:} For the GPT-4o generated portion, data was collected through API calls using Python scripts. The human-authored content was gathered via online Google Sheets containing the target homograph words and detailed instructions, as documented in our methodology. No additional validation procedures were applied to these collection mechanisms.

\paragraph{If the dataset is a sample from a larger set, what was
    the sampling strategy (e.g., deterministic, probabilistic with
    specific sampling probabilities)?}

\paragraph{ANS:} The dataset incorporates: (1) complete samples from ManaTTS and GPTInformal (covering all available data at study time), and (2) a non-random subset of CommonVoice selected by availability (prioritizing validated samples while respecting original data ordering). GPT-4o generations and human annotations were collected specifically for this study.

\paragraph{Who was involved in the data collection process (e.g.,
    students, crowdworkers, contractors) and how were they compensated
    (e.g., how much were crowdworkers paid)?}

\paragraph{ANS:} The human-annotated portion of the dataset was collected through voluntary participation of native Persian speakers from diverse backgrounds. While we did not collect detailed demographic information about participants, their native language proficiency was the primary qualification for contribution. Participants were not financially compensated, as the data collection was conducted as part of an academic research initiative.

\paragraph{Over what timeframe was the data collected?} Does this
  timeframe match the creation timeframe of the data associated with
  the instances (e.g., recent crawl of old news articles)?  If not,
  please describe the timeframe in which the data associated with the
  instances was created.

\paragraph{ANS:} The dataset was compiled in 2024-2025, combining newly generated GPT-4o outputs and human annotations with existing public corpora. The ManaTTS, GPTInformal, and CommonVoice components originate from their 2024 releases.

\paragraph{Were any ethical review processes conducted (e.g., by an
    institutional review board)?} If so, please provide a description
  of these review processes, including the outcomes, as well as a link
  or other access point to any supporting documentation.

\paragraph{ANS:} No ethical review processes were conducted.

\paragraph{Did you collect the data from the individuals in
    question directly, or obtain it via third parties or other sources
    (e.g., websites)?}

\paragraph{ANS:} The data was obtained from the individuals directly.

\paragraph{Were the individuals in question notified about the data
    collection?} If so, please describe (or show with screenshots or
  other information) how notice was provided, and provide a link or
  other access point to, or otherwise reproduce, the exact language of
  the notification itself.

\paragraph{ANS:} The data was not collected from a pre-existing source; instead, individuals were explicitly instructed to generate the data, eliminating the need for notification.

\paragraph{Did the individuals in question consent to the
    collection and use of their data?} If so, please describe (or show
  with screenshots or other information) how consent was requested and
  provided, and provide a link or other access point to, or otherwise
  reproduce, the exact language to which the individuals consented.

\paragraph{ANS:} Similar to the previous response, since the data was generated based on explicit instructions provided to the individuals, consent was inherently obtained through participation, and no additional consent process was necessary.

\paragraph{If consent was obtained, were the consenting individuals
    provided with a mechanism to revoke their consent in the future or
    for certain uses?} If so, please provide a description, as well as
  a link or other access point to the mechanism (if appropriate).

\paragraph{ANS:} As the data generation was based on direct instructions and not from pre-existing sources or personal information, the issue of consent revocation does not apply in this context.

\paragraph{Has an analysis of the potential impact of the dataset
    and its use on data subjects (e.g., a data protection impact
    analysis) been conducted?} If so, please provide a description of
  this analysis, including the outcomes, as well as a link or other
  access point to any supporting documentation.

\paragraph{ANS:} No such analysis has been conducted.

\paragraph{Any other comments?}

\paragraph{ANS:} No.

\subsection{Preprocessing/cleaning/labeling}

The questions in this section are intended
to provide dataset consumers with the information they need to
determine whether the ``raw'' data has been processed in ways that are
compatible with their chosen tasks. For example, text that has been
converted into a ``bag-of-words'' is not suitable for tasks involving
word order.\\

\paragraph{Was any preprocessing/cleaning/labeling of the data done
    (e.g., discretization or bucketing, tokenization, part-of-speech
    tagging, SIFT feature extraction, removal of instances, processing
    of missing values)?} If so, please provide a description. If not,
  you may skip the remaining questions in this section.

\paragraph{ANS:} Yes, the underlying text corpora sourced from previous datasets and generated through GPT-4o or human annotators were phonemized as labels using the LLM prompting method outlined in a prior study, as referenced in the paper.

\paragraph{Was the ``raw'' data saved in addition to the preprocessed/cleaned/labeled data (e.g., to support unanticipated future uses)?} If so, please provide a link or other access point to the ``raw'' data.

\paragraph{ANS:} Yes, the raw data includes the underlying text corpora from previous datasets (ManaTTS, GPTInformal, CommonVoice), as well as data generated using GPT-4o and contributions from human subjects. These data remain accessible and were only augmented with the phoneme labels as described earlier.

\paragraph{Is the software that was used to preprocess/clean/label the data available?} If so, please provide a link or other access point.

\paragraph{ANS:} Yes, the complete code for data generation and labeling is publicly accessible at https://github.com/MahtaFetrat/HomoRich-G2P-Persian.

\paragraph{Any other comments?}

\paragraph{ANS:} No.

\subsection{Uses}

The questions in this section are intended to encourage dataset
creators to reflect on the tasks for which the dataset should and
should not be used. By explicitly highlighting these tasks, dataset
creators can help dataset consumers to make informed decisions,
thereby avoiding potential risks or harms.\\

\paragraph{Has the dataset been used for any tasks already?} If so, please provide a description.

\paragraph{ANS:} Yes, it has been employed to finetune two neural G2P models and enhance a rule-based G2P tool in our research, which is used to evaluate data efficiency.

\paragraph{Is there a repository that links to any or all papers or systems that use the dataset?} If so, please provide a link or other access point.

\paragraph{ANS:} 
The dataset was not publicly available before this work, and as far as we know, it hasn't been utilized in any other projects.

\paragraph{What (other) tasks could the dataset be used for?}

\paragraph{ANS:} The dataset can be utilized for both general G2P conversion and specific homograph pronunciation disambiguation. Additionally, it could be valuable for tasks involving context understanding, such as word sense disambiguation. While not all sense disambiguations involve pronunciation differences, words with multiple pronunciations often convey distinct meanings that require contextual clarification.

\paragraph{Is there anything about the composition of the dataset or the way it was collected and preprocessed/cleaned/labeled that might impact future uses?} For example, is there anything that a dataset consumer might need to know to avoid uses that could result in unfair treatment of individuals or groups (e.g., stereotyping, quality of service issues) or other risks or harms (e.g., legal risks, financial harms)? If so, please provide a description. Is there anything a dataset consumer could do to mitigate these risks or harms?

\paragraph{ANS:} We do not believe that the dataset carries such risks.

\paragraph{Are there tasks for which the dataset should not be used?} If so, please provide a description.

\paragraph{ANS:} We do not foresee any specific limitations regarding potential uses of the dataset.

\paragraph{Any other comments?}

\paragraph{ANS:} No.

\subsection{Distribution}

\paragraph{Will the dataset be distributed to third parties outside of the entity (e.g., company, institution, organization) on behalf of which the dataset was created?} If so, please provide a description.

\paragraph{ANS:} Yes, the dataset is available to the public under a CC-0 license.

\paragraph{How will the dataset be distributed (e.g., tarball on website, API, GitHub)?} Does the dataset have a digital object identifier (DOI)?

\paragraph{ANS:} The dataset is distributed through the Hugging Face platform and will be assigned a DOI upon finalization.

\paragraph{When will the dataset be distributed?}

\paragraph{ANS:} The dataset is publicly available at https://github.com/MahtaFetrat/HomoRich-G2P-Persian.

\paragraph{Will the dataset be distributed under a copyright or other intellectual property (IP) license, and/or under applicable terms of use (ToU)?} If so, please describe this license and/or ToU, and provide a link or other access point to, or otherwise reproduce, any relevant licensing terms or ToU, as well as any fees associated with these restrictions.

\paragraph{ANS:} The dataset is shared under the CC-0 license, allowing free use.

\paragraph{Have any third parties imposed IP-based or other restrictions on the data associated with the instances?} If so, please describe these restrictions, and provide a link or other access point to, or otherwise reproduce, any relevant licensing terms, as well as any fees associated with these restrictions.

\paragraph{ANS:} No, there are no IP-based or other restrictions imposed on the data associated with the instances.

\paragraph{Do any export controls or other regulatory restrictions apply to the dataset or to individual instances?} If so, please describe these restrictions, and provide a link or other access point to, or otherwise reproduce, any supporting documentation.

\paragraph{ANS:} No, there are no export controls or other regulatory restrictions applicable to the dataset or individual instances.

\paragraph{Any other comments?}

\paragraph{ANS:} No.

\subsection{Maintenance}

The questions in this section are intended to
encourage dataset creators to plan for dataset maintenance and
communicate this plan to dataset consumers.\\

\paragraph{Who will be supporting/hosting/maintaining the dataset?}

\paragraph{ANS:} The dataset is stored in public data repositories and maintained by the authors for updates.

\paragraph{How can the owner/curator/manager of the dataset be contacted (e.g., email address)?}

\paragraph{ANS:} 
You can contact the authors via the following email addresses:
\begin{itemize}
    \item Mahta Fetrat: m.fetrat@sharif.edu
    \item Zahra Dehghanian: zahra.dehghanian97@sharif.edu
    \item Hamid R. Rabiee: rabiee@sharif.edu
\end{itemize}

\paragraph{Is there an erratum?} If so, please provide a link or other access point.

\paragraph{ANS:} There is currently no erratum. 

\paragraph{Will the dataset be updated (e.g., to correct labeling
    errors, add new instances, delete instances)?} If so, please
  describe how often, by whom, and how updates will be communicated to
  dataset consumers (e.g., mailing list, GitHub)?

\paragraph{ANS:} We intend to update the dataset if significant errors are identified or if valuable community contributions can be incorporated. However, we do not plan to establish a formal mechanism for communicating changes. Updates can be tracked through the version history available on the hosting platforms (e.g., GitHub).

\paragraph{If the dataset relates to people, are there applicable
    limits on the retention of the data associated with the instances
    (e.g., were the individuals in question told that their data would
     be retained for a fixed period of time and then deleted)?} If so,
    please describe these limits and explain how they will be
    enforced.

\paragraph{ANS:} There are no retention limits specified for the dataset.

\paragraph{Will older versions of the dataset continue to be
    supported/hosted/maintained?} If so, please describe how. If not,
  please describe how its obsolescence will be communicated to dataset
  consumers.

\paragraph{ANS:} No, older versions will not be maintained. We do not plan to implement a specific mechanism to notify consumers of updates. Instead, changes can be observed through the version history available on the hosting platforms (e.g., GitHub).

\paragraph{If others want to extend/augment/build on/contribute to
    the dataset, is there a mechanism for them to do so?} If so,
  please provide a description. Will these contributions be
  validated/verified? If so, please describe how. If not, why not? Is
  there a process for communicating/distributing these contributions
  to dataset consumers? If so, please provide a description.

\paragraph{ANS:} Contributions are very welcome. Contributors can open issues or submit pull requests on GitHub, or contact the authors directly for error reports or improvements. 

\paragraph{Any other comments?}

\paragraph{ANS:} No.

\end{document}